\documentclass{article}

\usepackage{hyperref}       

\usepackage[preprint]{neurips_2025}

\usepackage{amsmath,amsfonts,bm}

\usepackage[nameinlink,capitalize]{cleveref}
\crefname{theorem}{Theorem}{Theorems}
\crefname{lemma}{Lemma}{Lemmas}
\crefname{corollary}{Corollary}{Corollaries}
\crefname{proposition}{Proposition}{Propositions}
\crefname{conjecture}{Conjecture}{Conjectures}
\crefname{claim}{Claim}{Claims}
\crefname{definition}{Definition}{Definitions}
\crefname{assumption}{Assumption}{Assumptions}
\crefname{example}{Example}{Examples}
\crefname{remark}{Remark}{Remarks}
\crefname{observation}{Observation}{Observations}
\crefname{equation}{Eq.}{Eqs.}

\newcommand{\mask}{\mathrm{MASK}}

\newcommand{\del}[1]{{\left( #1\right)}}
\newcommand{\curly}[1]{{\left\{ #1\right\}}}
\newcommand{\norm}[1]{{\left\lVert #1\right\rVert}}

\def\vec#1{\mathbf{#1}}

\def\eqref#1{equation~\ref{#1}}

\def\ceil#1{\lceil #1 \rceil}
\def\floor#1{\lfloor #1 \rfloor}
\def\1{\bm{1}}

\def\bin{\textrm{bin}}

\def\mP{{\bm{P}}}

\DeclareMathAlphabet{\mathsfit}{\encodingdefault}{\sfdefault}{m}{sl}
\SetMathAlphabet{\mathsfit}{bold}{\encodingdefault}{\sfdefault}{bx}{n}

\newcommand{\R}{\mathbb{R}}

\newcommand{\KL}{D_{\mathrm{KL}}}

\usepackage{wrapfig}
\usepackage{graphicx}
\usepackage{subcaption}
\usepackage{algorithm}
\usepackage{algpseudocode}
\usepackage{bbm}
\usepackage{calc} 

\newsavebox{\mygridbox}

\def\abs#1{\left|#1\right|}
\def\norm#1{\left\lVert#1\right\rVert}
\def\curly#1{\left\{#1\right\}}

\def\v#1{\boldsymbol{#1}}

\def\colorful{1}

\ifnum\colorful=1

\newcommand{\snote}[1]{\footnote{\textcolor{red}{[SK: #1]}}}

\else

\newcommand{\snote}[1]{}

\fi
\def\partialmask{1}

\ifnum\partialmask=1
\newcommand{\xunmasked}{\vec{x}^{-M}}
\else
\newcommand{\xunmasked}{\vec{x}}
\fi
\def\vec#1{\mathbf{#1}}

\usepackage[utf8]{inputenc} 
\usepackage[T1]{fontenc}    
\usepackage{url}            
\usepackage{booktabs}       
\usepackage{amsfonts}       
\usepackage{nicefrac}       
\usepackage{microtype}      
\usepackage{xcolor}         
\usepackage{todonotes}
\usepackage{thmtools}

\usepackage{amsthm}

\usepackage{todonotes}
\theoremstyle{plain}
\newtheorem{theorem}{Theorem}[section]

\theoremstyle{definition}
\newtheorem{definition}[theorem]{Definition}

\theoremstyle{remark}

\newtheorem*{theorem*}{Theorem}
\newtheorem*{lemma*}{Lemma}
\newtheorem*{corollary*}{Corollary}

\usepackage[nameinlink,capitalize]{cleveref}
\crefname{theorem}{Theorem}{Theorems}
\crefname{lemma}{Lemma}{Lemmas}
\crefname{corollary}{Corollary}{Corollaries}
\crefname{proposition}{Proposition}{Propositions}
\crefname{conjecture}{Conjecture}{Conjectures}
\crefname{claim}{Claim}{Claims}
\crefname{definition}{Definition}{Definitions}
\crefname{assumption}{Assumption}{Assumptions}
\crefname{example}{Example}{Examples}
\crefname{remark}{Remark}{Remarks}
\crefname{observation}{Observation}{Observations}
\crefname{equation}{Eq.}{Eqs.}

\title{Parallel Sampling from Masked Diffusion Models via Conditional Independence Testing}

\author{Iskander Azangulov 
\thanks{This research was performed while the author was an intern at Microsoft Research Cambridge.} \\
Department of Statistics\\
University of Oxford\\
\texttt{iskander.azangulov@spc.ox.ac.uk} \\
\And
Teodora Pandeva \\
Microsoft Research, Cambridge\\
\texttt{t-tepandeva@microsoft.com} \\
\And
Niranjani Prasad \\
Microsoft Research, Cambridge\\
\texttt{Niranjani.Prasad@microsoft.com} \\
\And
Javier Zazo \\
Microsoft Research, Cambridge\\
\texttt{javierzazo@microsoft.com@microsoft.com} \\
\And
Sushrut Karmalkar \\
Microsoft Research, Cambridge\\
\texttt{skarmalkar@microsoft.com} \\
}

\begin{document}

\maketitle

\begin{abstract}

Masked diffusion models (MDMs) offer a compelling alternative 
to autoregressive models (ARMs) for discrete text generation
because they enable parallel token sampling, rather than sequential, 
left-to-right generation.
This means potentially much faster inference.
However, effective parallel sampling faces two competing requirements:
(i) simultaneously updated tokens must be conditionally independent, and
(ii) updates should prioritise high-confidence predictions. 
These goals conflict because high-confidence predictions often cluster 
and depend on each other, opportunities for parallel updates.

We present PUNT, a model-agnostic sampler that reconciles this trade-off. 
Our method identifies token dependencies and removes
lower-confidence tokens from conflicting groups. 
This produces sets of indices for unmasking that satisfy both independence and 
confidence criteria. Our approach ensures improved parallel unmasking 
through approximate conditional independence testing.

Our experiments show that PUNT delivers 
a superior trade-off between accuracy and compute when compared to other strong training-free baselines,
especially for generation of longer sequences. On the IFEval benchmark, it achieves up to 16\% higher accuracy over baseline methods,
including sequential generation (one-by-one).
These gains hold across different values of hyperparameters, mitigating the need for brittle hyperparameter tuning.
Moreover, we observe that PUNT induces an emergent hierarchical generation strategy, 
where the model first establishes high-level paragraph structure before local refinement, suggesting a planning-like generation process that contributes to strong alignment performance.

\end{abstract}

\section{Introduction}
\label{sec:intro}

The widespread deployment of Large Language Models (LLMs) has created massive computational workloads, 
consuming significant datacenter resources and electricity, thereby incurring substantial operational costs.
A primary driver of this inefficiency is inference speed, which is bottlenecked by the sequential, 
left-to-right generation process inherent in standard autoregressive models.
To overcome this, alternative methods have been developed to enable multiple tokens to be generated simultaneously.

Among approaches with the potential for parallel decoding, 
Masked Diffusion Models (MDMs) 
have emerged as a particularly promising framework
\citep{austin2023structured,lou2024discreteratios,nie2025lldm}.
Unlike autoregressive models, 
MDMs iteratively refine masked sequences by predicting 
\emph{subsets} of positions simultaneously, enabling parallel decoding.
However, determining which tokens to unmask in parallel 
without degradation in quality remains challenging.

Various inference strategies have been proposed to accelerate MDMs, 
including 
confidence-based token selection 
\citep{sahoo2024simpleeffectivemaskeddiffusion,patel2025accelerated}, 
structured unmasking patterns 
\citep{luxembourg2025planspeeddilatedscheduling, arriola2025blockdiffusion}, 
remasking 
\citep{wang2025remaskingdiscretediffusionmodels}, 
and distillation \citep{zhu2025dimo}.
However, these approaches share a critical limitation: 
they do not explicitly test for inter-token interference during parallel decoding.
Structured patterns impose rigid, data-agnostic schedules that 
ignore sequence-specific dependencies, while 
remasking and distillation either add computational overhead or require expensive retraining.

\paragraph{Our Contribution.}

We propose a different approach to parallel decoding based on \emph{contextual independence}
—testing whether tokens can be decoded in parallel by checking 
for independence at the sampled point, 
rather than for all possible outcomes. 
Unlike standard conditional independence, 
which requires integrating over all possible outcomes
(which is computationally prohibitive for large token spaces), 
contextual independence provides
the part that matters at the current decoding step.

To find the contextually independent subsets, 
we propose PUNT (Parallel Unmasking with Non-influence Tests), 
a training-free procedure that employs a divide-and-conquer strategy. 
Our algorithm selects ``anchor'' subsets and tests entire ``candidate'' 
groups for dependence in batch.
By carefully designing splits, 
PUNT certifies a large block of tokens for parallel generation using 
only $O(\log m)$ model calls per step (compared with $m$ for fully sequential unmasking) where $m$ is the number of masked tokens. 

    PUNT enjoys the following advantages:
    first, 
    PUNT is \textbf{training-free} and requires no model fine-tuning or distillation.
    Second, 
unlike rigid structured patterns or confidence-based approaches, 
PUNT \textbf{dynamically adapts} to sequence-specific dependencies.
For instance, we see that it exhibits an emergent \textbf{hierarchical generation} strategy, where
the model first establishes high-level paragraph structure before refining the details.
Third,
for long-form text generation on alignment benchmarks as well as \emph{de novo} protein generation tasks, 
PUNT outperforms other standard baselines and quickly reaches 
its maximum quality with very few forward passes when compared to 
other algorithms, resulting in a \textbf{stable Pareto frontier} over the number of forward evaluations of the MDM.

\textbf{Organization.} 
Section~\ref{sec:background} introduces masked diffusion models and formalizes the parallel decoding problem.
Section~\ref{sec:method} presents our main algorithmic contribution,
Section~\ref{sec:experiments} presents empirical evaluation, and
Section~\ref{sec:related} discusses related work. 
Finally, Section~\ref{sec:conclusion} discusses implications and future directions.

\section{Background on Masked Diffusion Models}
\label{sec:background}

In this section, we review the fundamentals of masked diffusion models and establish the notation used throughout this paper.

\textbf{Notation.}
We denote vectors with bold lowercase (e.g., $\vec{x}$) 
and scalars with regular lowercase (e.g., $y$); 
random variables use uppercase (e.g., $X, \vec{Y}$) 
with corresponding lowercase for their realizations (e.g., $x, \vec{y}$). 
We will also use uppercase letters to denote sets and tensors, when it is clear from context. 
Let $L = [\ell]:=\curly{1,\ldots, \ell}$ denote integers up to $\ell$. 
For $I \subseteq L$, $-I:=L\setminus I$ is its complement, 
and $\vec{x}^I = \{x^i \mid i \in I\}$ represents the indexed subset of sequence 
$\vec{x} = (x^1,\ldots, x^\ell)$. 
With vocabulary $V$, we consider discrete state space $V^{L}$. 
For random sequence $\vec{X} = (X^1, \dots, X^\ell)$, we write 
$p(\vec{x}) := \mP(\vec{X} = \vec{x})$ for outcome 
$\vec{x} \in V^{L}$, and $p^j(\cdot)$ for the marginal at position $j$. 
Extending to $V_{\mask} = V \cup \{\mask\}$, 
a token $x^i$ is \textbf{masked} if $x^i = \mask$. 
For any sequence, $M = M(\vec{x}) := \{i \mid x^i = \mask\}$ denotes masked indices, 
with unmasked indices denoted $-M$. 
Conditional distribution at position $j$ given observed tokens $\vec{x}^I$ 
is written as $p^j(\cdot \mid \vec{x}^I)$, 
shorthand for $\mP(X^j = \cdot \mid \vec{X}^I = \vec{x}^I)$. 
For a sequence with masked indices $M$, $\xunmasked$ denotes unmasked tokens. 
In the iterative generation process, $\vec{x}_t$ represents the sequence at step $t$. 
and $M_t := M(\vec{x}_t)$ denotes masked indices at step $t$.

\textbf{Masked Language Modeling.}
Masked language modeling trains neural networks to predict missing tokens from context. 
Given a sequence $\vec{x} \in V_{\mask}^\ell$ with masked positions $M$, 
the model parameterized by $\theta$ learns conditional distributions:
\[
p_\theta^i(\cdot \mid \xunmasked) \approx p^i(\cdot \mid \xunmasked), \quad \forall i \in M.
\]
During training, 
a clean sample (i.e. without any masked coordinates) 
$\vec{x}_{\text{clean}} \sim p(\cdot)$ is drawn from the true data distribution. 
A random subset of its tokens $M \subseteq L$ is then selected to be masked, 
creating the corrupted sequence $\vec{x}$ 
where $\xunmasked = \vec{x}_{\text{clean}}^{-M}$ 
and $\vec{x}^M$ consists of $\mask$ tokens.

The model parameters $\theta$ are optimized to maximize the conditional 
log-likelihood of the original tokens at masked positions:
\[
\mathcal{L}(\theta) = \mathbb{E}_{\vec{x}_{\text{clean}}, M} \left[ \sum_{i \in M} \log p_\theta^i(x^i_{\text{clean}} \mid \xunmasked) \right].
\]
Generation proceeds iteratively using the trained model.
Starting from a fully masked sequence 
$\vec{x}_0 = (\mask, \dots, \mask)$, each iteration performs two operations at timestep $t$:
(1) sample candidate tokens 
$y_t^i \sim p_\theta^i(\cdot \mid \vec{x}^{-M_t}_t)$ 
for all masked positions $i \in M_t$, and
(2) update a subset $R \subseteq M_t$ of these positions with their sampled values, 
so that $x_{t+1}^i \leftarrow y_t^i$ for $i \in R$ and $x_{t+1}^i \leftarrow x_t^i$ for $i \notin R$.
This process repeats until all positions are unmasked, 
producing the final sequence $\vec{x}_T$.

The iterative sampling process introduces 
two key sources of error at each step that can compromise sample quality. 
For clarity, we analyze the error within a single step and drop the time index $t$ 
in the following discussion.

\textbf{Approximation Error.} 
The learned model $p_\theta^i(\cdot \mid \xunmasked)$ only 
\emph{approximates} the true conditional distribution 
$p^i(\cdot \mid \xunmasked)$. This potentially leads to suboptimal token predictions. 
To mitigate this, we employ a confidence score 
$\phi_i$ per position $i$, to guide mask selection at each step, 
updating only those positions where the model exhibits high confidence. 
This strategy improves generation quality by prioritizing high-confidence predictions. Common confidence scores $\phi_i$ for position $i$ include:
\textbf{negative entropy} $\sum_{x\in V} p_\theta^i(x \mid \xunmasked) \log p_\theta^i(x \mid \xunmasked)$,
\textbf{confidence}
$p_\theta^i(y^i \mid \xunmasked)$ of the sampled token, and
\textbf{top margin} between the two most likely tokens.

\textbf{Joint Dependencies.} 
A more fundamental limitation arises from the sampling strategy itself. 
When sampling masked tokens $y^i$ independently from their conditional distributions $p_\theta^i(\cdot \mid \xunmasked)$, 
we implicitly assume conditional independence among all masked tokens given the unmasked context. 
Natural sequences, however, exhibit complex dependencies that violate this assumption. 
True conditional independence for candidate tokens $R \subseteq M$ requires the joint probability to factorize as:
\begin{equation}
\label{eq:ind}
p^R(\cdot \mid \xunmasked) = \prod_{i \in R} p^i(\cdot \mid \xunmasked)
\end{equation}
In general, the joint distribution does not factorize in this way. Finding a subset of tokens where this factorization holds presents significant computational challenges, as verifying this condition requires checking that \eqref{eq:ind} holds for all outcomes 
$\vec{y}^R\in V^R$—a space that grows exponentially with $|R|$. 
The following section presents an efficient method to identify token subsets that approximately satisfy this independence condition, 
thereby mitigating this source of error.

\section{Method}
\label{sec:method}
In this section, we introduce our method for one step of parallel unmasking. 
We first establish \textbf{contextual independence} 
as the criterion for safe parallel unmasking (\S\ref{sec:contextual_independence}), 
then present our \textbf{efficient subset discovery} algorithm that identifies independent 
token sets using only $O(\log|M|)$ model evaluations (\S\ref{sec:efficient_discovery}).

\subsection{Contextual Independence}
\label{sec:contextual_independence}
To address joint dependencies, we adopt the notion of \textit{contextual independence} 
as our criterion for parallel unmasking. 
This property precisely characterizes when parallel sampling yields the 
same distribution as sequential sampling. 
Unlike full statistical independence (overly restrictive) 
or confidence-based heuristics (which ignore dependencies), 
contextual independence identifies tokens that can be unmasked simultaneously 
given the current context.

\begin{definition}[Contextually Independent Random Variables]
A random variable $X$ is \emph{contextually independent} 
of a random variable $Y$ at a point $y$ if the conditional distribution of $X$ given $Y=y$ 
is identical to the marginal distribution of $X$, i.e., $p_{X \mid Y}(\cdot \mid Y=y) = p_X(\cdot)$.
\end{definition}

\begin{definition}[Contextually Independent Sequences]
A \emph{sequence} of random variables $(X^1, \dots, X^\ell)$ is 
\emph{contextually independent} at an outcome $(x^1, \dots, x^\ell)$, 
if for each $i \in L$, the conditional distribution of $X^i$ 
given the preceding outcomes $\vec{x}_{<i} = (x^1, \dots, x^{i-1})$ 
is identical to its marginal distribution. Formally, for all $i \in L$:
$p_{X^i \mid \vec{X}_{<i}}(\cdot \mid \vec{x}_{<i}) = p_{X^i}(\cdot)$, 
where $\vec{X}_{<i} = (X^1, \dots, X^{i-1})$.
\end{definition}

In other words, under the contextual independence assumption, 
sampling the vector $(x^1, \dots, x^\ell)$ 
sequentially is equivalent to sampling its components in parallel.

Our goal is, given a candidate vector $\vec{y}^M$, 
to find an \emph{ordered}
\footnote{Eventually we will be ordering these via confidence metrics $\phi_i$, 
see Section~\ref{sec:implementation}.}
set of masked indices 
$R = \{r_1, \dots, r_{|R|}\} \subseteq M$ that 
are contextually independent 
relative to the unmasked context $\xunmasked$. 
Formally, for any $i \in \{1, \dots, |R|\}$, 
the distribution at position $r_i$ must be independent 
of the outcomes at preceding positions $R_{<i} := \{r_j \mid j < i\}$, 
given the unmasked context:
\begin{equation}
\label{eq:context_indep_goal}
p^{r_i}(\cdot \mid \xunmasked, \vec{y}^{R_{<i}}) = p^{r_i}(\cdot \mid \xunmasked).
\end{equation}

A naive, greedy approach to construct such a set would be to iterate through 
all masked indices $m \in M$ and sequentially add an index to $R$ 
if it satisfies Equation~\ref{eq:context_indep_goal} given the previously added indices. 
However, this requires $O(|M|)$ sequential model evaluations, 
which defeats the purpose of parallel sampling.

We propose an efficient recursive algorithm based on a recursive divide-and-conquer strategy, 
which we will later provide an an efficient iterative implementation for.
The validity of this approach relies on the following stability assumption regarding 
the conditional independence structure of the model.

\begin{restatable}{assumption}{IndependenceStability}(Independence Stability)
\label{asmp:1}
Let $i \in M$ be a masked index, and let  
$U\subseteq M \setminus \{i\}$ be a subset of masked indices. 
If for some sequence of tokens $\vec{y}^U$ we have 
$p^i(\cdot \mid \vec{y}^U, \xunmasked) = p^i(\cdot \mid \xunmasked),$
then for any $W\subset U$ it holds that 
$p^i(\cdot \mid \vec{y}^W, \xunmasked) = p^i(\cdot \mid \xunmasked).$
\end{restatable}

This assumption represents a balanced compromise 
between complete independence and simple contextual independence.
It states that if a set of positions $U$ does not influence the prediction at position $i$, 
then any subset $W\subset U$ will also not influence that prediction.
This property ensures that independence tests conducted at any stage 
of our recursive algorithm remain valid throughout all subsequent stages.
Section~\ref{sec:independence_stability} provides a justification for why this assumption 
is reasonable for transformer-based architectures 
and empirical evidence that it approximately holds in practice.

\begin{figure}
    \centering
\includegraphics[width=1.0\linewidth]{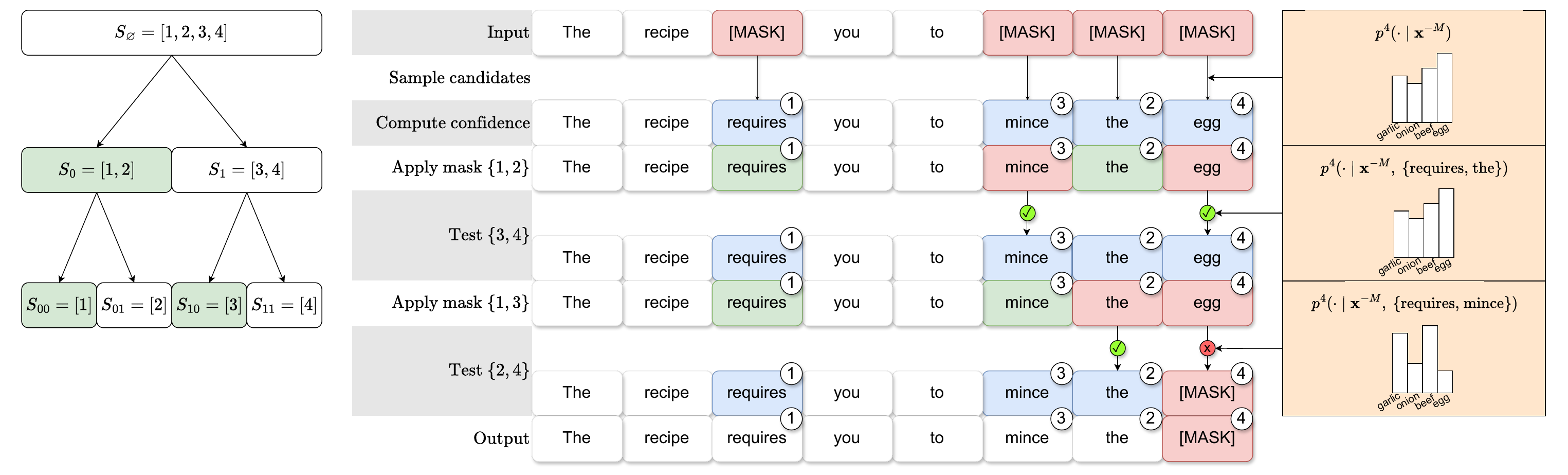}
    \caption{
    Illustration of one iteration of PUNT on 4 masked tokens, 
    which consists of $\log_2 4 = 2$ tests.
    Left: A binary tree representing the recursive partitioning. 
    Each level corresponds to a single parallel test in the iterative algorithm.
    Right: In each round, 
    confidence-ordered tokens (circled numbers) 
    are partitioned into ``anchor'' (green) and ``test'' sets. 
    Test tokens that are dependent on the anchor set are rejected (red), 
    while independent ones (blue) are kept.}
    \label{fig:punt_illustration}
\end{figure}

\subsection{Efficient Subset Discovery}
\label{sec:efficient_discovery}

Under Assumption~\ref{asmp:1}, 
we can construct the set $R$ in $O(\log|M|)$ parallel steps. 
If there is at least one masked position (i.e. $|M| \geq 1$), 
the recursive algorithm starts with $S=M$ and proceeds as follows 
(see Figure~\ref{fig:punt_illustration} for an illustration of one iteration):

At each recursive call, 
its input is a (confidence)
\footnote{See \Cref{sec:implementation} for details} 
ordered subset of masked candidates $S = (s_1, s_2, \ldots, s_{|S|})\subseteq M$. 
The base case for the recursion is when $|S| \le 1$, 
in which case the procedure returns $S$. 
For larger sets, the algorithm proceeds as follows:

\begin{enumerate}
    \item \textbf{Divide:} 
    The ordered input set $S$ is split into two balanced halves: the ``anchor" set 
    $S_0 = (s_1, \ldots, s_{p})$ 
    and the ``test" set 
    $S_1 = (s_{p+1}, \ldots, s_{|S|})$,
    where $p$ is a split point of the designer's choice. 

    \item \textbf{Prune} (Filter)\textbf{:} 
    The ``test" set $S_1$ is pruned based on its dependency on the candidates $\vec{y}^{S_0}$. 
    For each index $i \in S_1$, we compute its new conditional distribution 
    and measure the change from the baseline using the KL divergence: 
    \[
    \varepsilon_i := \KL \del{ p^i(\cdot \mid \xunmasked, \vec{y}^{S_0}) \, \big\| \, p^i(\cdot \mid \xunmasked) }
    \]
    A filtered set $S'_1$ is then formed 
    by retaining only those indices for which the change is below a threshold $\varepsilon > 0$: $ S'_1 = \{i \in S_1 \mid \varepsilon_i < \varepsilon\}.$

    \item \textbf{Recurse:} 
    The algorithm then makes two independent (parallel) recursive calls: 
    one on the ``anchor" set $S_0$ and 
    another on the filtered ``test" set $S'_1$ and obtains $R_0$ and $R_1$ respectively. 

    \item \textbf{Combine:} 
    The final result $R$ for the input set $S$ is the \textbf{union} (ordered sum) 
    of the outputs $R:=R_0\sqcup R_1$ from the two recursive calls above. 
    Note that by construction any token in $R_1$ is contextually independent of $S_0$ 
    and by Assumption~\ref{asmp:1}, 
    it is contextually independent of subset $R_0 \subset S_0$. 
\end{enumerate}

Choosing $p= \lfloor |S|/2 \rfloor$ for each recursive iteration, 
we ensure that the recursion depth is  $O(\log|M|)$. 
In the next section, 
we discuss how to execute all calls at the same recursion level 
using a single network evaluation, 
thereby achieving $O(\log|M|)$ cost per round.

\subsection{Confidence Alignment and Implementation Details}
\label{sec:implementation}

This section addresses two key implementation aspects: 
(i) incorporating confidence-based prioritization into our recursive algorithm 
to maintain generation quality, and 
(ii) transforming the recursive procedure into an efficient iterative implementation.

\textbf{Confidence-Ordered Splits.}  
At each recursive step, 
the candidate set $S$ is split into $S_0$ and $S_1$, 
and positions in $S_1$ are pruned if they exhibit strong dependence on $S_0$. 
By sorting the initial candidate set $S$ in descending order of confidence 
(see Section~\ref{sec:background} for options),
$\phi_{s_1} > \phi_{s_2} > \dots > \phi_{s_{|S|}}$, 
we ensure that $S_0$ always contains tokens with at least median-level confidence. 
Consequently, 
tokens pruned from $S_1$ necessarily have lower confidence than those retained in $S_0$.
In fact, this also ensures that during each unmasking step, 
the highest-confidence token in $M$ is always included in the final set $R$, 
since it will never be pruned.    

\textbf{Binary Encoding of Recursive Calls.}
To enable parallel computation, 
we transform our recursive algorithm into an efficient iterative procedure. 
At each level of the recursion tree, 
we combine all independence tests into a single, parallel model evaluation. 

To see how this might be done, 
suppose at some recursion level we test pairs $(S_0^1,S_1^1), (S_0^2,S_1^2),\ldots (S_0^k,S_1^k)$, 
by construction, these sets form a partition of a subset of $M$. 
We propose, instead of performing these tests independently, 
to test all ``test" tokens against the union of all ``anchor" sets $\bigsqcup_\ell S^\ell_0$.
Then, Assumption~\ref{asmp:1} ensures that passing this combined test implies passing the 
individual tests. Formally, for any $\ell$ and any $i\in S^\ell_1$ if 
$p^i_\theta(\cdot\mid \vec{y}^{\bigsqcup S^\ell_0}, \xunmasked) = p^i_\theta(\cdot\mid \xunmasked),$
then it also satisfies
$p^i_\theta(\cdot\mid \vec{y}^{S^\ell_0}, \xunmasked) = p^i_\theta(\cdot\mid \xunmasked).$

\emph{Binary Representation of Recursive Splits.}

We now describe how this idea can be employed 
to convert our recursive algorithm to an iterative one. 
The recursive splits are determined by each token's position in the confidence-ordered list $M$. 
This allows us to pre-determine all splits by assigning a binary code to each position.
More precisely, we assign each position $i \in \{1, \ldots, |M|\}$ a binary representation 
$\text{bin}(i)$ with $\lceil \log_2 |M| \rceil$ bits, padded with zeros if necessary. 
Tracking this binary encoding allows us to identify the path of each node in the recursion tree.

At recursion level $b$, we have $2^b$ nodes, with each node indexed by a unique binary prefix
$\vec{u} \in \{0,1\}^b$ corresponding to the subset 
$S_{\vec{u}} = \curly{i\le |M|: \operatorname{prefix}_b[\bin(i)] = \vec{u}}$,
which is then partitioned into ``anchor" ($S_{\vec{u}0}$) and ``test" ($S_{\vec{u}1}$) subsets
(see Figure~\ref{fig:punt_illustration} (Left) for an example).

We would like to combine all $2^b$ tests at recursion level $b$ into a single test.
To do so, we define a global partition for level $b$ based on the $b$-th bit of the binary encoding.
$B_b = \{ i \in [|M|] : \text{the } b\text{-th bit of } \text{bin}(i) = 0 \}.$

Starting with $R = M$, 
each round $b$ can now partition the current set using the predefined binary split $B_b$: 
the ``anchor" tokens ($S_0 = R \cap B_b$) and ``test" tokens ($S_1 = R \setminus B_b$). 
All tokens in $S_1$ are tested for dependence on $\vec{y}^{S_0}$ 
in \emph{a single forward pass}, dependent tokens are removed from $R$. 
After all $\log |M|$ rounds complete, 
the remaining set $R$ contains only contextually independent tokens.

The resulting procedure, summarized below 
(and in Algorithm~\ref{alg:planner}), 
requires only $O(\log |M|)$ forward evaluations of the model per denoising step and guarantees 
that the returned set $R$ consists of contextually independent, 
high-confidence tokens that can be unmasked in parallel.

\emph{Iterative Algorithm.}
Given confidence-ordered masked tokens $M = \{m_1, m_2, \ldots, m_{|M|}\}$ 
where $\phi_{m_1} \geq \phi_{m_2} \geq \cdots \geq \phi_{m_{|M|}}$, 
we initialize $R \leftarrow M$ and execute $\lceil \log_2 |M| \rceil$ iterations.

For each iteration $b \in \{1, \ldots, \lceil \log_2 |M| \rceil\}$:
\begin{enumerate}
    \item \textbf{Test:} 
    Partition $R$ into anchor tokens $S_0 = R \cap B_b$ and test tokens $S_1 = R \setminus B_b$.
    \item \textbf{Prune:} For each $j \in S_1$, 
    compute the KL divergence 
    $d_j = D_{\text{KL}}(p^j(\cdot \mid \xunmasked) \,\|\, p^j(\cdot \mid \xunmasked, \vec{y}^{S_0}))$ 
    in a single forward pass.
    \item \textbf{Update:} 
    Remove dependent tokens from 
    $R$: $R \leftarrow R \setminus \{j \in S_1 : d_j > \varepsilon\}$.
\end{enumerate}

The final set $R$ contains tokens that can be unmasked in parallel without interfering with each other.

\begin{figure}[t]
    \centering
    \begin{subfigure}{0.45\textwidth}
        \centering
        \includegraphics[scale=0.19]{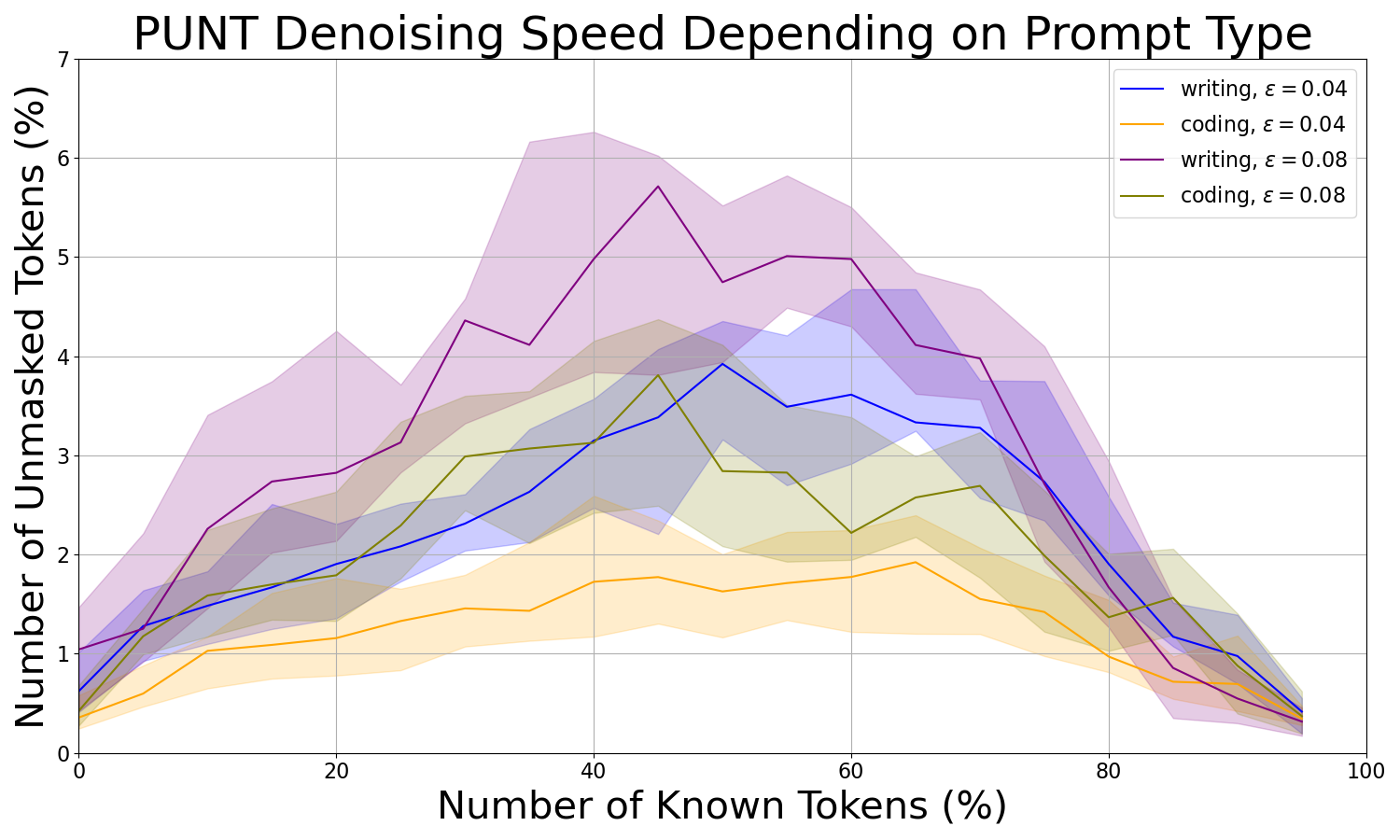}
    \end{subfigure}
    \hfill
    \begin{subfigure}{0.45\textwidth}
        \centering
        \tiny{Step 9}
        \includegraphics[scale=0.15]{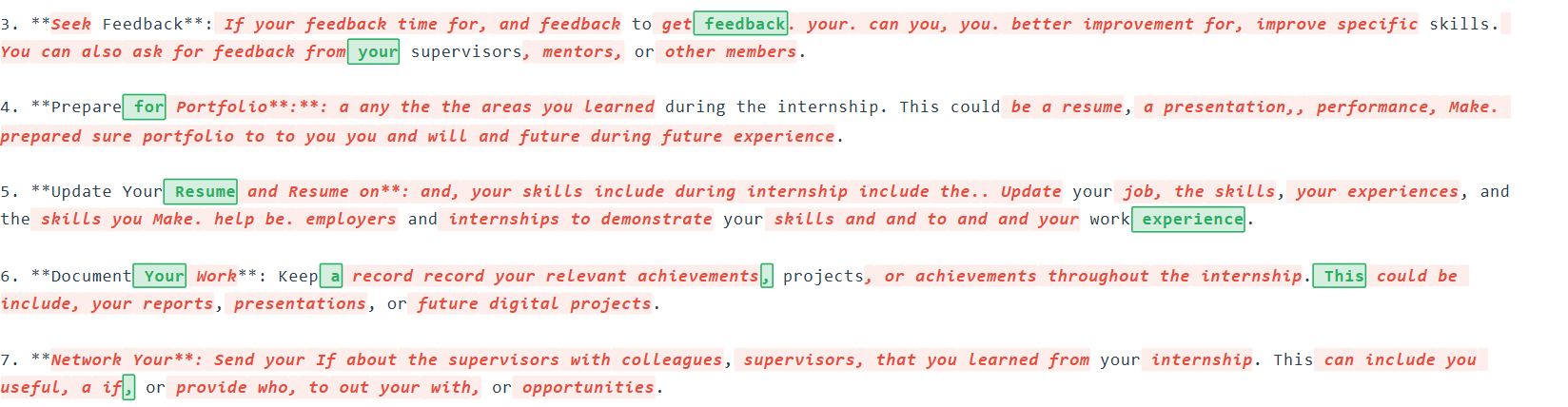}
        \tiny{Step 18}
        \includegraphics[scale=0.15]{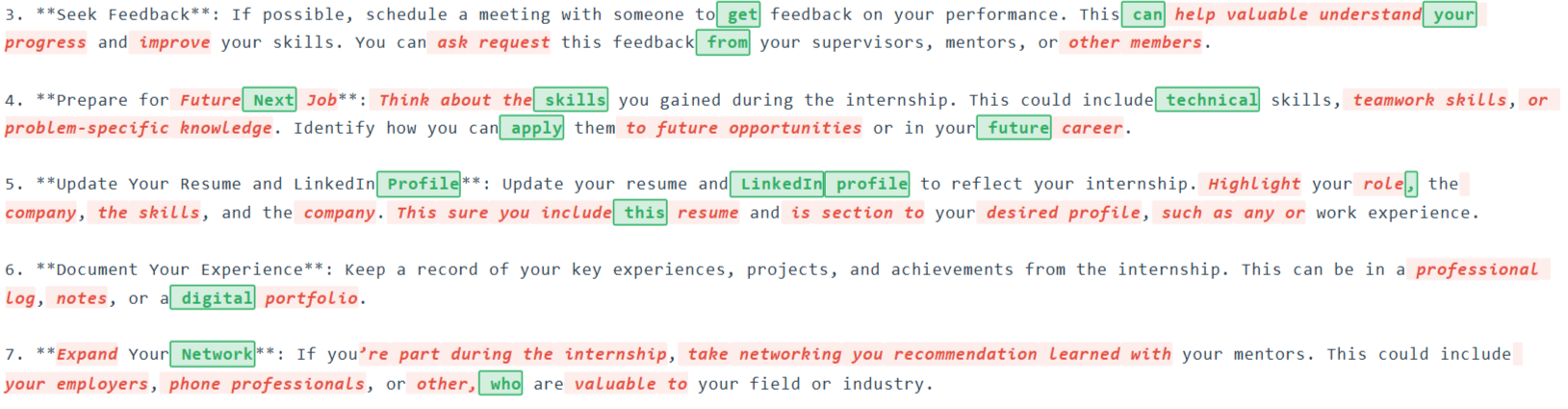}
        \label{fig:img3}
    \end{subfigure}
    \caption{
        Left: Unmasking efficiency for various prompt types vs number of unmasked tokens. 
        Right: Visualization of the denoising process at steps 9 and 18 for a sample prompt
        (``What should I do at the end of the internship.").
        Tokens are color-coded: green tokens are accepted by PUNT for parallel unmasking in the current step, 
        red tokens are rejected, and uncolored tokens were unmasked in previous steps.
        See \cref{sec:unmaskings} for more examples of intermediate denoising steps.}
    \label{fig:sampler_properties}
\end{figure}

\subsection{Algorithmic Properties and Independence Stability}

This section discusses PUNT's properties for text generation and justifies Assumption~\ref{asmp:1}.

\textbf{Adaptive Unmasking.}
Our sampler exhibits emergent hierarchical generation, 
first establishing high-level structure (e.g., paragraphs, headings) before filling in details, 
as shown in~\cref{fig:sampler_properties} (right).
As can be seen, at step 9, the model has already generated 
the main headings and subheadings of the article, while the rest of the text remains masked. 
By step 18, the model has begun filling in the details under each heading.

We hypothesize this behavior stems from the conditional independence 
between high-level structural tokens and fine-grained details. 
Because the latter exert minimal influence on the former, 
the structural tokens pass the independence tests in PUNT's filtration stage and are unmasked early. 
A formal investigation of this phenomenon is left for future work.

This hierarchical generation has a cascading effect. 
Once revealed, high-level tokens act as contextual anchors, 
partitioning the text into conditionally independent sections. 
This allows the sampler to unmask tokens in different sections in parallel, 
adapting its denoising speed to the task's inherent structure. 
As shown in~\cref{fig:sampler_properties} (left), 
this results in different performance profiles for various prompts. 

\textbf{Independence Stability of Transformers.}
Assumption~\ref{asmp:1} (Independence Stability) is a direct consequence 
of the Transformer architecture's attention mechanism. 
In Transformers, the influence of one token on another is governed by attention weights; 
if the attention from position $i$ to position $j$ is zero, 
then position $j$ has no direct influence on the representation at position $i$.

This relationship between attention and influence allows us to connect conditional independence 
to attention scores. Specifically, we argue that a token $y_i$ is conditionally independent 
of a set of tokens $\vec{y}^V$ given the remaining tokens $\xunmasked$ 
if and only if the total attention from position $i$ to all positions in $R$ is negligible 
across all layers and heads, i.e.,
\begin{center}
\textit{
The conditional distribution $p_\theta^i(\cdot \mid \vec{y}^R, \xunmasked)$ 
equals $p_\theta^i(\cdot \mid \xunmasked)$ 
if and only if\footnote{Except for padding end-of-sequence(EOS) tokens, see details in Appendix~\ref{sec:independence_stability}} 
the cumulative attention weight from position $i$ to all positions in $R$ is negligible.
}
\end{center}
This property directly implies Independence Stability. 
Since attention weights are non-negative, 
if the attention from position $i$ to a set $R$ is negligible, 
the attention to any subset $U \subset R$ must also be negligible. 
A detailed justification is provided in Appendix~\ref{sec:independence_stability}.

\section{Experiments}
\label{sec:experiments}

\begin{figure}
    \centering
    \begin{subfigure}{0.48\textwidth}
        \centering
        \includegraphics[width=1.05\linewidth]{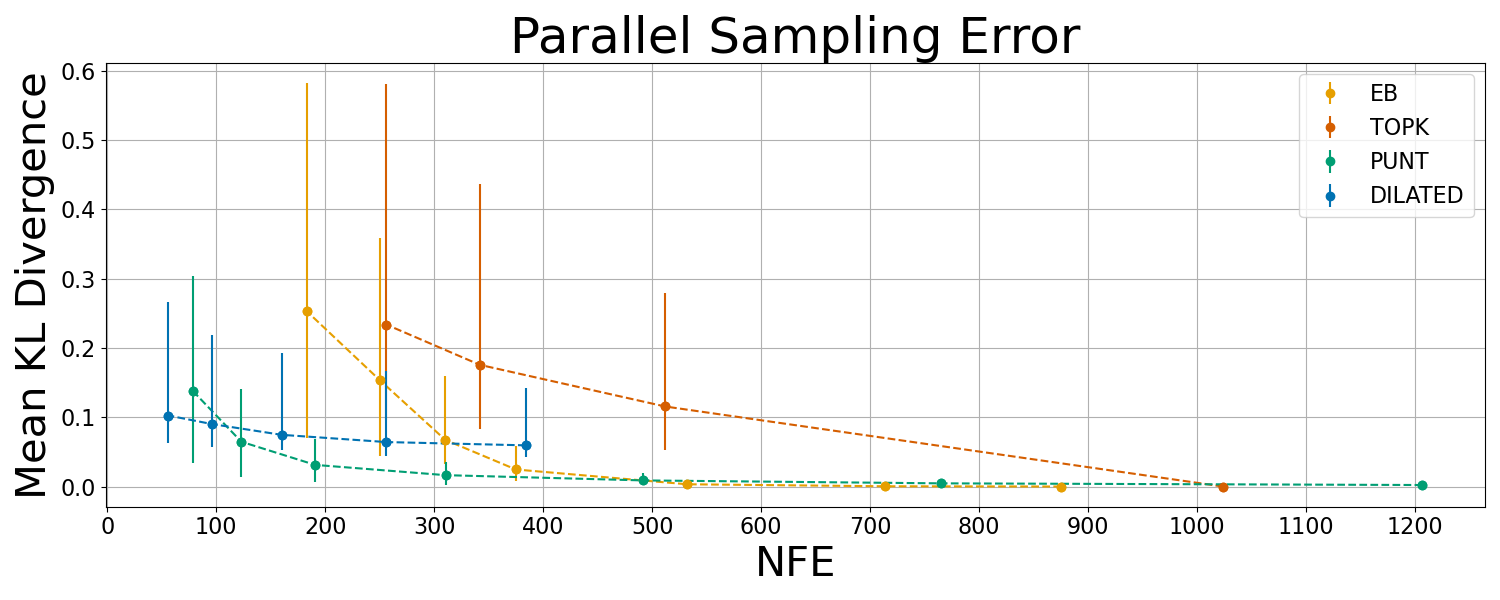}
        \label{fig:sub1}
    \end{subfigure}
    \hfill
    \begin{subfigure}{0.48\textwidth}
        \centering
        \includegraphics[width=1.05\linewidth]{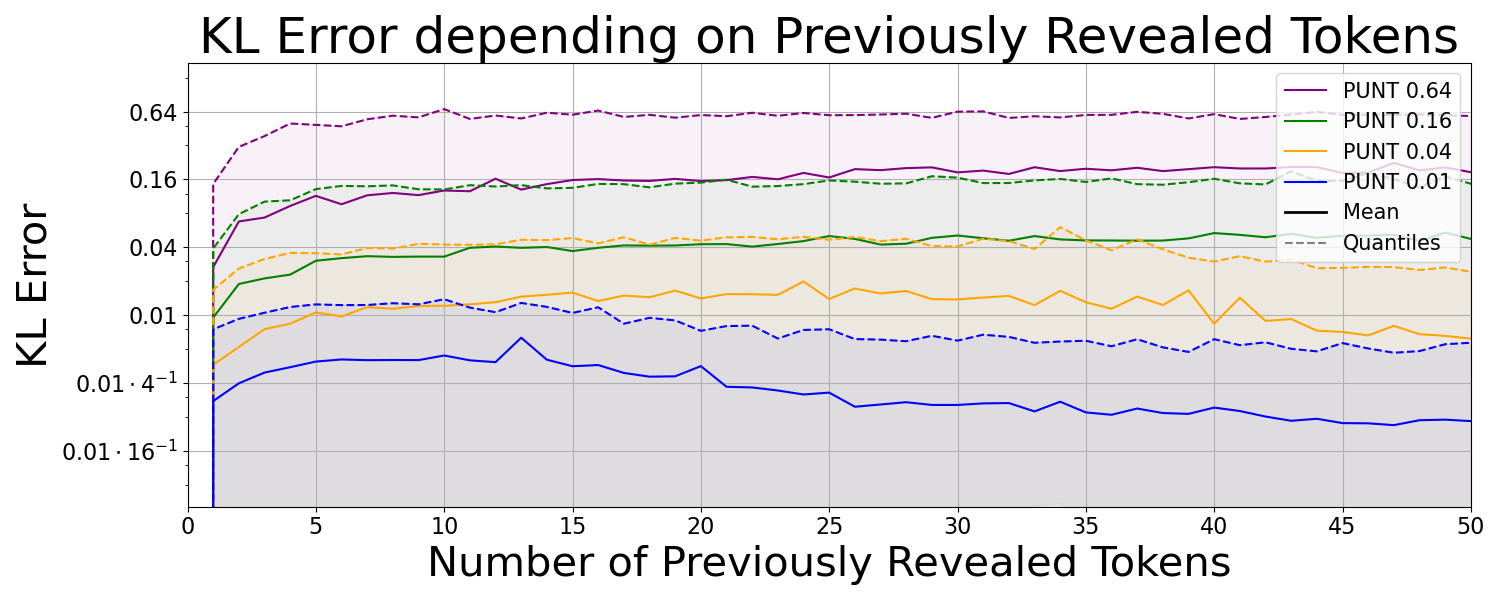}
        \label{fig:sub2}
    \end{subfigure}
    \caption{Parallel sampling error~(\eqref{eq: kl_error}). Left: Average error for different samplers compared to number of forward evaluations (NFEs). Right: Median together with confidence intervals $(Q_{5}, Q_{95})$ of the error for PUNT samplers with different $\varepsilon$ as a function of the number of previously revealed tokens. Note that $Q_5$ remains below $10^{-3}$ across all positions. }
    \label{fig: kl_error}
\end{figure}

We evaluate our proposed sampler, PUNT, on a number of natural language tasks. 
Our empirical results validate the effectiveness of our approach 
and support the attention hypothesis introduced in \cref{sec:independence_stability}.
We evaluate:
(i) PUNT's performance on long-form text generation tasks such as MTBench;
(ii) PUNT's effectiveness on short-answer benchmarks for mathematics and code generation;
(iii) The error introduced by parallel token sampling and its relationship to the exploration rate $\varepsilon$;
(iv) Deviation in empirical attention patterns (supporting theoretical independence assumptions).

\textbf{Experimental Setup.}
We evaluate PUNT on two powerful, open-source large language models: 
Dream 7B \citep{ye2025dream} and LLaDA 1.5 \citep{zhu2025llada}.  
 
\textbf{Baselines.}
We compare PUNT against three strong, 
training-free baseline samplers. These baselines include: 
(i) standard $\mathrm{top}$-$k$ sampling; 
(ii) the EB-sampler \citep{patel2025accelerated}; and 
(iii) the Dilated-sampler \citep{luxembourg2025planspeeddilatedscheduling}. 
Note that all the samplers are implemented for a given context length $|M|$, 
in a \textbf{non-semi-autoregressive} manner and the exact parameter configurations are 
provided in~\cref{subsec:results}.

We also evaluate the performance of PUNT on generation in a structured biological domain. Specifically, we look at unconditional generation of \emph{de novo} membrane proteins using MemDLM~\citep{goel2024memdlm}, a state-of-the-art protein language model. Full details are provided in~\cref{appdx:prot_experiments}.

\subsection{Alignment Benchmarks}

\begin{figure}[t!]
    \centering
    \includegraphics[width=1.05\columnwidth]{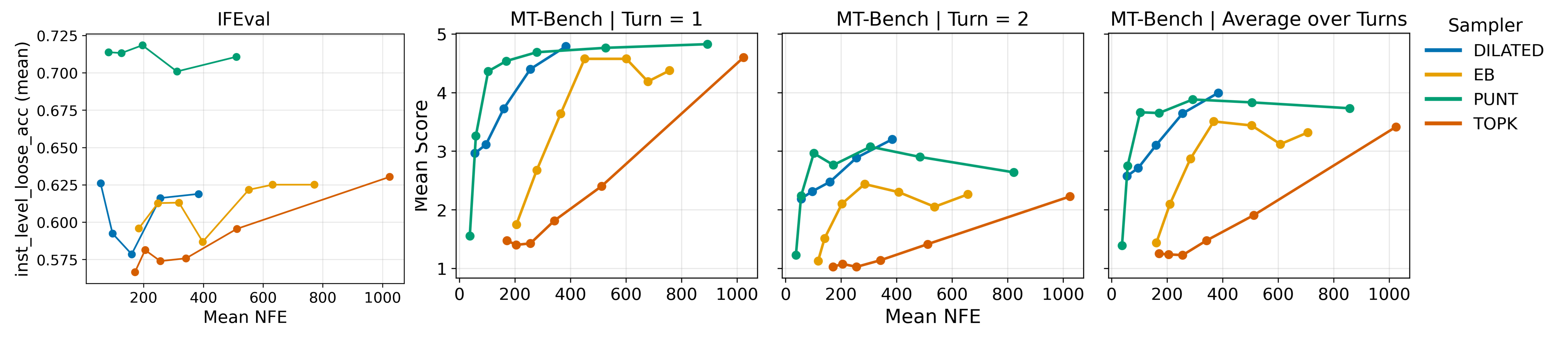}
    \caption{IFEval and MTBench performance of PUNT compared to baselines on Dream 7B. Benchmark specific scores (higher is better) vs mean number of forward passes.}
    \label{fig:ifeval-mtbench}
\end{figure}

We evaluate PUNT on the instruction-following benchmarks 
MTBench~\citep{bai2024mt} and IFEval~\citep{zhou2023instructionfollowingevaluationlargelanguage}. 
MTBench comprises 80 tasks across diverse domains—including 
creative writing, logical reasoning, and code generation—providing a robust evaluation framework for model performance. 
Each task consists of two turns, with the second turn depending on the output of the first. 
IFEval complements this by specifically measuring instruction-following accuracy through 
a collection of carefully designed test cases that evaluate precise adherence to complex instructions. 
Full benchmark details are provided in \Cref{subsec:setup}. 
As shown in Figure~\ref{fig:ifeval-mtbench}, 
PUNT consistently outperforms all baseline samplers on both benchmarks, 
achieving higher scores across both metrics (\texttt{inst\_level\_loose\_acc} 
and mean score; higher is better).
 Importantly, it delivers these gains while requiring substantially fewer forward evaluations (NFEs).

\subsection{Short-answer Benchmarks}

\begin{wrapfigure}{r}{0.5\textwidth}
    \centering
    \includegraphics[width=0.48\textwidth]{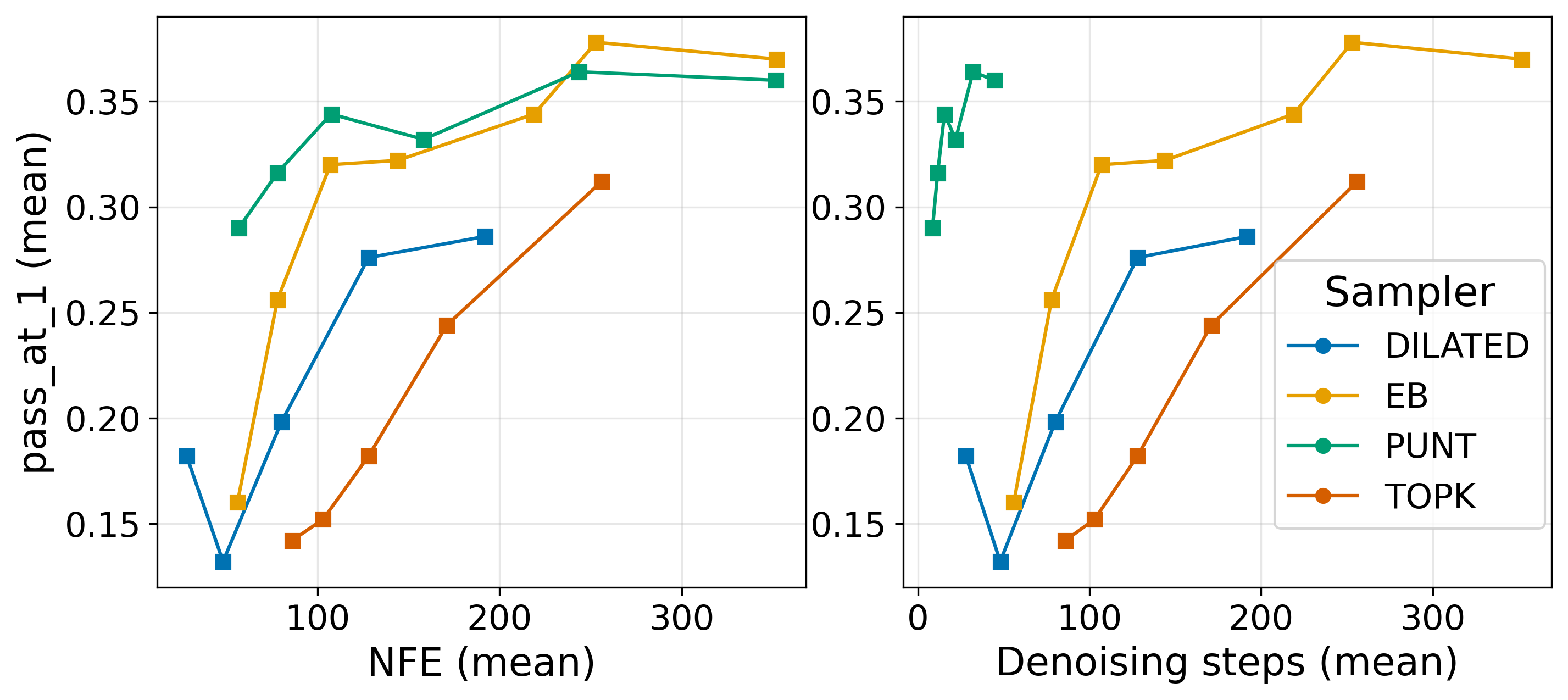}
    \caption{MBPP performance of PUNT compared to baselines on Llada. MBPP pass@1 (higher is better) vs mean number of forward passes.}
    \label{fig:mpbb}
\end{wrapfigure}

We evaluate PUNT across diverse benchmarks spanning mathematical reasoning 
(GSM8K~\citep{cobbe2021gsm8k}, 
and code generation 
(HumanEval~\citep{chen2021humaneval},
MBPP~\citep{austin2021mbpp}).

As expected, PUNT underperforms on short-answer tasks with limited context since 
it requires multiple forward passes for complete generation.  
For instance, in the MBPP benchmark on LLaDA with temperature 0.7 (see \cref{fig:mpbb}), 
PUNT's performance aligns closely with EB when evaluated by the number of forward evaluations (NFE). 
However, when measured by the number of denoising steps—that is, the number of algorithmic steps, 
where each step may involve multiple forward passes in parallel—PUNT outperforms other samplers.
A simple fix could be to use PUNT only for the latter part of the generation,
but we leave this for future work. 
Similar trends are observed across other benchmarks in this group.  
Results comparing all samplers, models, and hyperparameters appear in~\cref{subsec:results}.

\subsection{PUNT Sampler Error Analysis}
\label{sec: sampler error analysis}
We empirically quantify the parallel sampling error on the LLaDA model. 
For this, we generate 1024-token responses for MTBench~\citep{bai2024mt} 
with exploration rates $\varepsilon \in \{0.01, \dots, 0.32\}$. 
Within each parallel generation step, tokens are ordered by confidence before sampling. 
For the $i$-th token at position $r_i$ unmasked in parallel, 
we compute the error between the true conditional distribution and our independence approximation:
\begin{equation} 
\label{eq: kl_error}
\delta^{r_i}_{\operatorname{KL}}= \KL\del{p_\theta^{r_i}(\cdot \mid \xunmasked, \vec{y}^{R_{<i}}) \,\big\|\, p_\theta^{r_i}(\cdot \mid \xunmasked)}.
\end{equation}
This KL divergence quantifies the information lost by assuming token $r_i$ is conditionally independent 
of other tokens unmasked in the same step, $\vec{y}^{R_{<i}}$.
As shown in \cref{fig: kl_error}, 
PUNT achieves a low parallel sampling error while maintaining a small NFE compared to other samplers.
Furthermore, the parallel decoding error, $\delta_{\operatorname{KL}}$, 
remains robustly below the $\varepsilon$ threshold, 
irrespective of the number of tokens previously revealed in the step.

\label{sec: experiment_sampling_error}

\section{Related work}
\label{sec:related}

Our work builds on recent advances in discrete diffusion models and 
inference-time planners. 

\textbf{Masked Diffusion Models (MDMs).}
Discrete diffusion models \citep{austin2023structured} 
offer a non-autoregressive alternative for text generation. 
Training objectives based on score matching \citep{lou2024discreteratios} 
and masked language modeling \citep{sahoo2024simpleeffectivemaskeddiffusion} 
have enabled large-scale models like LLaDA \citep{nie2025lldm} and others \citep{nie2025scalingmdm}. 
Commercial implementations include 
Gemini Diffusion \citep{deepmind2025geminidiffusion} and Mercury \citep{inceptionlabs2025mercury}.
MDMs face two key limitations: 
compounding errors from parallel unmasking and inefficient KV caching. 
We address the former by identifying token sets for safe parallel unmasking, 
which minimizes interference and improves both efficiency and quality.

\textbf{Inference-Time Planners for Acceleration.}
Efficient inference scheduling remains MDMs' central challenge. 
Various training-free planners aim to minimize function evaluations (NFEs) 
while maintaining generation quality.

\emph{Confidence and Entropy Gating.} 
Confidence-based scheduling iteratively unmasks tokens with highest model confidence (or lowest entropy) 
\citep{sahoo2024simpleeffectivemaskeddiffusion}. 
The EB-Sampler extends this by dynamically unmasking 
variable-sized token sets whose aggregate entropy stays below threshold 
$\gamma$ \citep{patel2025accelerated}. 
While adaptive, these methods remain conservative, 
ignore token independence, and typically unmask only small subsets.

\emph{Remasking and Refinement.} 
Several methods correct parallel decoding errors through remasking.
ReMDM \citep{wang2025remaskingdiscretediffusionmodels} iteratively remasks and updates generated tokens.
Path-Planning (P2) \citep{peng2025path} and DDPD \citep{liu2025ddpd} separate inference into planning (selecting tokens to update/remask) and denoising stages.
While improving quality, these approaches increase NFE through corrective passes.

\emph{Spacing Schedulers.} 
These fixed-geometry (non-adaptive) methods enforce spatial separation between parallel unmaskings.
Dilated scheduling unmasks non-adjacent token groups for improved stability
\citep{luxembourg2025planspeeddilatedscheduling}. 
Halton-based schedulers use low-discrepancy sequences for uniform spacing
\citep{besnier2025haltonschedulermaskedgenerative}. 
Block Diffusion balances AR and parallel generation by processing contiguous spans
\citep{arriola2025blockdiffusion}. 

\emph{Analysis of Ordering and Scheduling.} 
Recent theoretical and empirical work
has deepened the community's understanding of these schedulers. \citet{kim2025train}
study the impact of token ordering, showing that adaptive inference can sidestep
computationally hard subproblems. \citet{park2024jump} focus on optimizing the temporal
schedule (the number and placement of diffusion steps) to reduce NFEs. Others have
explored MDLMs for complex reasoning, where planning is critical \citep{ye2025beyondautoregression},
and for specialized domains like code generation \citep{gong2025diffucoder}.

\textbf{Comparison to Autoregressive Accelerators.}
While autoregressive models like LLaMA-3 \citep{grattafiori2024llama3herd} 
are accelerated by speculative decoding \citep{leviathan2023speculative,xia2022speculative}, 
this approach remains fundamentally sequential. 
In contrast, our method reduces NFEs by leveraging the non-sequential, 
any-order generation capabilities of MDMs. 
Orthogonal optimizations like KV caching are applicable to both paradigms 
\citep{ma2025dkvcache,hu2025kv}.

\section{Conclusion and Future Work}
\label{sec:conclusion}

We introduced \textsc{PUNT}, 
a training-free sampler that looks to resolve the conflict between speed and quality in MDMs by efficiently identifying sets of approximately 
conditionally independent tokens for parallel unmasking. 
This enables a significant reduction in the number of model evaluations needed for generation while preserving output quality.
We provided a conceptual justification for its applicability to transformer architectures and validated its effectiveness on mathematics, code, and long‑form text benchmarks. 
We also observe that PUNT induces an emergent hierarchical generation strategy: coarse paragraph structure is established early, followed by localized refinement.

Future work can extend this approach in several directions: (i) developing adaptive or curriculum-style schedules for the independence threshold $\epsilon$ to balance early exploration with late precision; (ii) distilling PUNT into a student model that predicts contextually independent reveal sets in a single forward pass; and (iii) combining PUNT with orthogonal efficiency techniques such as KV-caching, to further shift the accuracy–compute Pareto frontier.

\paragraph{Acknowledgments}{
We gratefully acknowledge Grigory Bartosh for valuable discussions and Ted Meeds and Finale Doshi-Velez for helpful comments on the manuscript.
}
\bibliography{references}
\bibliographystyle{apalike}

\newpage
\appendix
\begin{center}
    {\LARGE\bfseries Appendix}
\end{center}


\section{Organisation of Appendix}
\label{sec:organisation}

The rest of the appendix is organized as follows.
In \cref{sec:independence_stability}, 
we justify Assumption~\ref{asmp:1} by demonstrating that it holds 
for Transformer-based masked language models,
which is a direct consequence of the Transformer's attention mechanism.
In \cref{appdx:experiments}, we provide additional experimental details and results.
In \cref{appdx:prot_experiments}, we provide some preliminary experiments on a protein masked diffusion model.
In \cref{sec:unmaskings}, we provide two examples of text that is generated by PUNT.

\emph{Remark on Notation:}
In addition to standard notation as defined in the paper, 
in the appendix, we will also use upper-case bold letters (such as $\textbf{A}$) to denote tensors. 
We will use lowercase and unbolded letters to denote scalars (such as $A_{ij}$).
In addition, we may have uppercase letters (such as $Q, K, V$) annotations to help annotate different matrices.
This is to accommodate standard notation used in the literature. 

\section{Independence Stability}
\label{sec:independence_stability}

In this section, we demonstrate that Assumption~\ref{asmp:1} 
holds for Transformer-based masked language models, 
which is a direct consequence of the Transformer's attention mechanism. 
Let us start with recalling the assumption. 
\IndependenceStability*
Next, we recall the design of attention mechanism and discuss prior works

\paragraph{Attention-Based Independence.}

In transformers~\cite{vaswani2023attentionneed}, 
the attention weights control information flow between positions.
For an input sequence $\vec{X} = (\vec{X}^1,\ldots, \vec{X}^L)\in \R^{L\times d_{in}}$, 
each attention head computes query, key, and value vectors for every position:
\[
\vec{Q}^i = \vec{W}^Q\vec{X}^i, \quad \vec{K}^i = \vec{W}^K\vec{X}^i, \quad \vec{V}^i = \vec{W}^V\vec{X}^i,
\]
where $\vec{W}^Q, \vec{W}^K \in \R^{d_k \times d_{in}}$ 
and $\vec{W}^V \in \R^{d_v \times d_{in}}$ are learned weight matrices.

The attention mechanism then computes pairwise attention scores between 
all positions through scaled dot products:
\[
\vec{A}=\mathrm{softmax}\!\left( \frac{\vec{Q}\vec{K}^\top}{\sqrt{d_k}} \right) \in \R^{L\times L},
\]
where $\vec{Q}, \vec{K} \in \R^{L \times d_k}$ stack the query and key vectors across all positions. 
The attention weights $A^{ij}$ quantify how much position $j$ influences position $i$, 
computed via normalized dot-product similarity. 
The output of one head combines value vectors weighted by these attention scores:
$\vec{head}_h =\vec{A} \vec{V} \in \R^{L\times d_v}.$
Finally, outputs of different heads are stacked to get,
$\vec{Z_i} = \textrm{concat}\del{\vec{head}^i_1,\ldots, \vec{head}^i_H}.$
The output of the layer $\vec{Y} = (\vec{Y}^1,\ldots, \vec{Y}^L)\in \R^{L\times d_{out}}$ 
is calculated as $\vec{Y}^i = F(\vec{Z}^i)$ 
by application to each of the coordinates of MLP together 
with normalization layers and skip connections. 

Crucially, the attention weights $A_{ij}$ directly control information flow: 
when $A_{ij} = 0$, position $j$'s value vector 
$\vec{V}^j$ contributes nothing to position $i$'s output. 
The model's final predictions are obtained by applying softmax to the last layer's output: 
$p_\theta^i(\cdot\mid \xunmasked) := \mathrm{softmax}(\vec{Y}^i)$. 
Therefore, $p_\theta^i(\cdot\mid \xunmasked) = p_\theta^i(\cdot\mid \vec{y}^R, \xunmasked)$ 
holds if and only if $\vec{Y}^i$ remains unchanged when tokens at positions $R$ are revealed.

\paragraph{Stability of Unmasked Tokens.} Recent works~\cite{hu2025kv, ma2025dkvcache} 
have demonstrated that during iterative inference, 
the query, key, and value vectors ($\vec{Q}^{-M}$, $\vec{K}^{-M}$, $\vec{V}^{-M}$) 
for already unmasked tokens, remain stable and can be cached for computational efficiency.

\begin{figure}
    \centering
    \includegraphics[width=\linewidth]{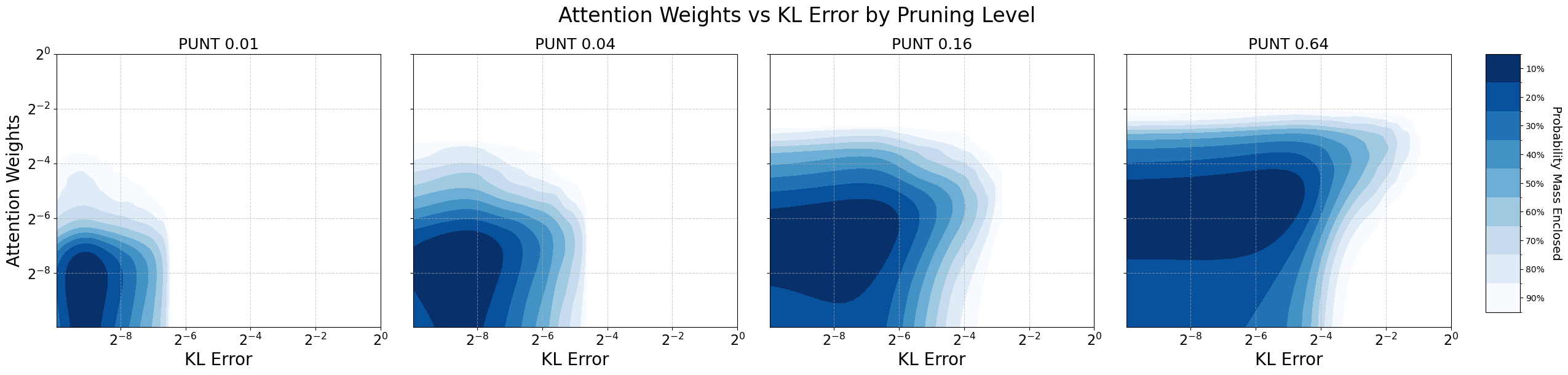}
    \caption{Joint distribution of $\delta_{\operatorname{KL}}$ -- the sampling error and $\delta_A$ -- the total attention to the previous tokens revealed in parallel.}
    \label{fig:attn_dist_kl_error}
\end{figure}
\paragraph{Why Independence Stability Holds.}

\begin{figure}
    \centering
    \includegraphics[width=\linewidth]{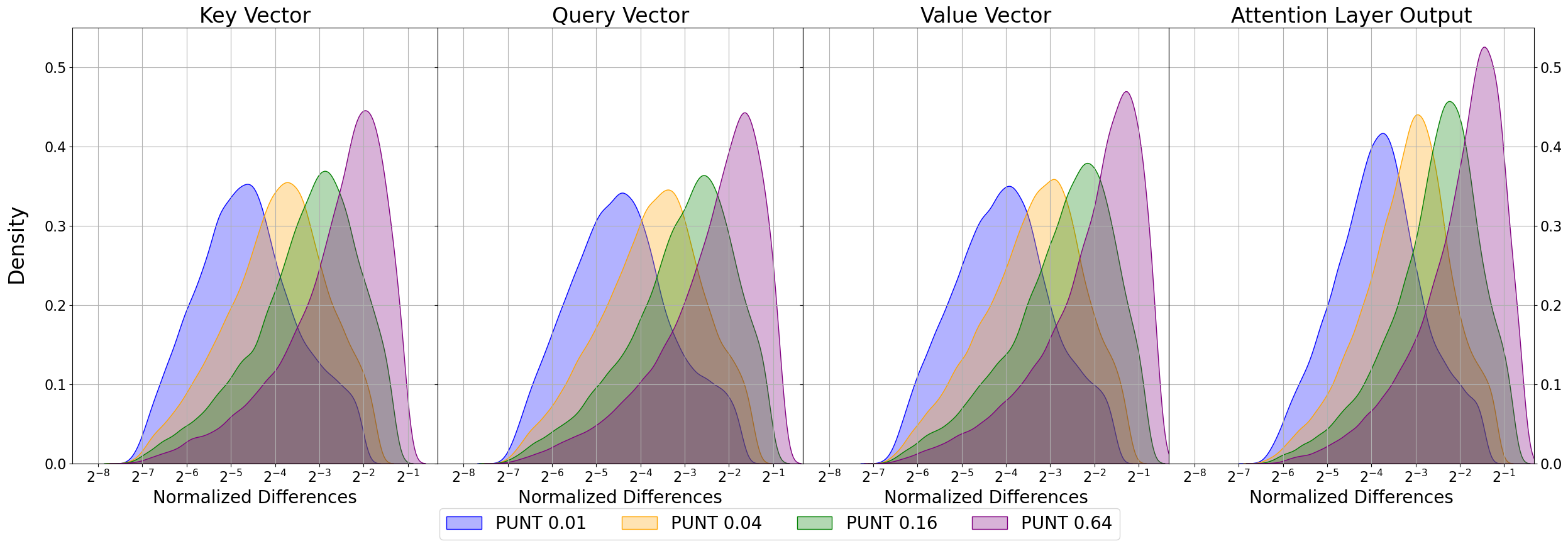}
    \caption{Difference between attention statistics for 
    $p^{r_i}(\cdot \mid \xunmasked, \vec{y}^{R_{<i}})$ 
    relative to the same statistics computed while evaluating $p^{r_i}(\cdot \mid \xunmasked)$.
    }
    \label{fig:attn}
\end{figure}

Let us return to Assumption~\ref{asmp:1}. First, we discuss padding end-of-sequence (EOS) tokens, which are used to fill the unused suffix reserved for an answer. By design, if there is an EOS token in $x^{-M}$ to the left of position $i$ then $p^i(\text{EOS}|x^{-M}) = 1$ and the assumption automatically holds.

For regular tokens, we note that the stability property implies that in both cases, when we condition on $(\vec{y}^U, \xunmasked)$ or $(\xunmasked)$, the representations  ($\vec{Q}^{-M}$, $\vec{K}^{-M}$, $\vec{V}^{-M}$) stay the same, while the main change happens for tokens in $U$. 

The stability property allows us to concentrate on the information flow between position $i$ and tentatively unmasked subset $\vec{y}^U$, which we recall governed by attention weights vector $\vec{A}_{iU} := \del{A_{i,u_1},\ldots, A_{i,u_{|U|}}}$.  Specifically, we argue that a token $y_i$ is conditionally independent 
of a set of tokens $\vec{y}^U$ given the remaining tokens $\xunmasked$ 
if and only if the total attention from position $i$ to all positions in $U$ is negligible 
across all layers and heads, or more formally, if $\norm{\vec{A}_{iU}}_1 = \sum_{u \in U} A_{iu} < \delta$ for some small $\delta > 0$.

Now consider any subset $W \subset U$. 
The non-negativity of attention weights (a direct consequence of the softmax operation) 
yields the inequality: 
\[
\norm{\vec{A}_{iW}}_1 = \sum_{w \in W} A_{iw} \leq \sum_{u \in U} A_{iu} = \norm{\vec{A}_{iU}}_1< \delta
\]

This demonstrates that if position $i$ pays negligible attention to the entire set $U$, 
it necessarily pays negligible attention to any subset $W \subset U$. 
Consequently, the conditional distribution at position $i$ remains approximately invariant when 
conditioning on tokens at positions in $W$: 
$p_\theta^i(\cdot \mid \vec{y}^W, \xunmasked) \approx p_\theta^i(\cdot \mid \xunmasked)$. 
This relationship directly corresponds to Assumption~\ref{asmp:1} (Independence Stability).

\paragraph{Empirical Validation}

We use the same setup as was used in~\cref{sec: sampler error analysis}, as a source of prompts, we use the first round requests from MTBench, and sample the responses using the PUNT algorithm with different thresholds $\varepsilon =\curly{0.01, 0.04, 0.16, 0.64}$. 

For a step of the PUNT sampler with threshold $\varepsilon$, let $R$ denote the set of tokens unmasked at this step, sorted according to the confidence, $\vec{y}^{R}$ denotes the set of sampled candidates, and $x^{-M}$ denotes the set of already revealed tokens.

As we demonstrated at \cref{fig: kl_error} sampled tokens satisfy
\[
\delta_{\operatorname{KL}}=\KL\del{p^{r_i}(\cdot \mid \xunmasked, \vec{y}^{R_{<i}}) \,\big \|\,  p^{r_i}(\cdot \mid \xunmasked)} < \varepsilon.
\]
For each token $r_i$, we compute the total attention from token $r_i$ to previously revealed tokens $R_{<i}$ for all heads of the last layer, i.e.
\[
\delta_{A} = \norm{\vec{A}_{r_iR_{<i}}}_1 = \sum_{j < i}A_{r_ir_j}
\]
and plot~(\cref{fig:attn_dist_kl_error}) the distribution of pairs ($\delta_{\operatorname{KL}},\delta_{A}$) for different thresholds. 

We also compute the change of the layer output $\vec{Y}^{r_i}$ and how it changes when we condition on $\vec{y}^{R_{<i}}$. We use the normalized difference metric to compute the change, which is defined as 
$
\operatorname{normalized\_difference}(a,b) := \norm{a-b}_2/\norm{a}_2,
$ and plot~(\cref{fig:attn}) the distribution of the change.
Finally, similar to unmasked tokens, we observed that representations $\vec{Q}^{r_i}$, $\vec{K}^{r_i}$, $\vec{V}^{r_i}$  of masked token $r_i$ in the attention layer also stays stable when we additionally condition on previously revealed tokens $\vec{y}^{R_{<i}}$.

\begin{algorithm}
\caption{PUNT (Parallel Unmasking with Non-influence Tests)}
\label{alg:planner}
\begin{algorithmic}[1]
\State \textbf{Input:} 
masked sequence $\vec{x}$, 
vector of candidates $\vec{y}$, 
threshold $\varepsilon$
\State \textbf{Output:} 
certified set $R\subseteq M$ to unmask in parallel
\State Sort masked indices w.r.t. confidence heuristic $\phi$ in decreasing order
\State Construct $M$, the set of all masked indices.
\State $R\gets M$
\State Let $B_b := \{ i \in [|M|] : \text{the } b\text{-th bit of } \text{bin}(i) = 0 \}.$
\For{$b$ in $[\log |M|]$}
    \State $S_0\gets R\cap B_b$; \hfill (positions to tentatively unmask)
    \State $S_1\gets R\setminus B_b$; \hfill (positions to check for dependence)
    \For{each $j\in S_1$}
        \State $d_j\gets \mathrm{D}_{\mathrm{KL}}\big(p^j(\cdot\mid\vec{x}^{-M})\,\big\|\,p^j(\cdot \mid\vec{x}^{-M},\vec{y}^{S_1})\big)$
        \If{$d_j>\varepsilon$} \State $R\gets R\setminus\{j\}$ \EndIf
    \EndFor
\EndFor
\State \Return $R$
\end{algorithmic}
\end{algorithm}

\section{Implementation and Experiments}\label{appdx:experiments}

This section evaluates the proposed planner PUNT (Algorithm~\ref{alg:planner}) 
across diverse sequence generation tasks. 
All experiments are conducted on A100 GPUs with 40GB memory.

PUNT offers a clear win in step efficiency without compromising on quality. 
However, this is not indicative of the underlying compute used, which is better captured by the number of forward passes (NFE).
In terms of NFE, it performs competitively, and particularly on long-sequence tasks, 
it often surpasses the baselines. 
We leave further per-step optimisation for future work. 

\subsection{Experimental Setup}
\label{subsec:setup}
We evaluate two state-of-the-art discrete diffusion models for natural language: 
\textbf{LLaDA-1.5}~\citep{zhu2025llada} and \textbf{Dream-v0-Instruct-7B}~\citep{ye2025dream} 
(referred to as Llada and Dream, respectively).  In this section, we detail the experimental setup, 
including tasks, datasets, evaluation metrics, and baseline methods.

\subsubsection*{Tasks and Datasets}
We assess PUNT's performance on a variety of sequence generation tasks. 
The evaluation relies on the following standard public datasets and their corresponding protocols:

\begin{itemize}
  \item Math word problems and formal math: GSM8K \citep{cobbe2021gsm8k}, MATH \citep{hendrycks2021math}
  \item Code generation: HumanEval \citep{chen2021humaneval} and MBPP \citep{austin2021mbpp}. 
  \item Instruction-following evaluation: IFEval \citep{zhou2023instructionfollowingevaluationlargelanguage}
  \item Open-ended question benchmarks: MT-Bench \citep{zheng2023judging}.
\end{itemize}

\subsubsection*{Evaluation Metrics and Configuration}
We use task-specific evaluation metrics and measure efficiency 
in terms of the number of forward evaluations and the number of iterations PUNT takes.

\textbf{Quality Metrics:}
\begin{itemize}
  \item Math problems: Match accuracy (GSM8K)
  \item Code generation: Pass@1 success rate (HumanEval, MBPP)
  \item Instruction following: Strict/Loose prompt/instruction adherence (IFEval)
  \item Open-ended generation: GPT-4o scoring 1-10 (MT-Bench)
\end{itemize}

\textbf{Efficiency Metrics:}
\begin{itemize}
  \item Number of network function evaluations (NFE) per sequence
  \item Number of generation steps (PUNT-specific)
\end{itemize}

\begin{table}[t]
\centering
\small
\begin{tabular}{l l r r r r r l}
\toprule
Experiment & NumFewshot & max length  \\
\midrule
GSM8K      &     4     & 512  \\
HumanEval  & 0 & 512    \\
MBPP   & 3 & 512  \\
IFEval & 0 & 1024 \\
MT-BENCH & - & 1024 \\
\bottomrule
\end{tabular}
\caption{Experimental configuration for each benchmark task.}
\label{tab:config}
\end{table}

\subsubsection*{Baseline Methods}
We compare against representative training-free schedulers with the following parameters:
\begin{itemize}
  \item Top-k Sampler with $k=1,2,3,4,5,6;$
  \item EB-Sampler (entropy-bounded unmasking) with ${\epsilon=0.01, 0.05, 0.1, 0.5, 1.0, 2.0, 4.0}$ \citep{patel2025accelerated};
  \item Geometry-aware spacing: dilated with log window size in $\{3,4,5,6,7\}$ \citep{luxembourg2025planspeeddilatedscheduling}, 
\end{itemize}

Each of these baselines utilizes a confidence score to rank positions by certainty. 
Different options for the confidence score are described below.

\subsubsection*{Confidence Scoring Strategies}

All confidence scoring strategies operate on the model's output probability distribution. 
For each position $t$ in a sequence, the model produces logits $l_{b,t,v}$ for every token $v$ in the vocabulary. 
These are converted into a probability distribution using the softmax function:
\[
p_{b,t,v} = \frac{e^{l_{b,t,v}}}{\sum_{v'=1}^V e^{l_{b,t,v'}}}.
\]
From this distribution, we compute a scalar confidence score $s_{b,t}$ that quantifies 
the model's certainty at that position. 
A higher score indicates greater confidence, prioritizing that position for earlier unmasking.
To define the scoring strategies, we use the following notation:
\begin{itemize}
  \item $p_{b,t,(k)}$: The $k$-th largest probability at position $t$, 
  such that $p_{b,t,(1)} \ge p_{b,t,(2)} \ge \dots \ge p_{b,t,(V)}$.
  \item $y_{b,t}$: The token actually sampled at position $t$.
\end{itemize}

\paragraph{Negative Entropy}
\[
s_{b,t} = \sum_{v=1}^V p_{b,t,v}\,\log p_{b,t,v} = - H\!\left(p_{b,t}\right),
\qquad
H\!\left(p_{b,t}\right) = - \sum_{v=1}^V p_{b,t,v}\,\log p_{b,t,v}.
\]
This is the \emph{negative} Shannon entropy. 
Values lie in $\big[-\log V,\,0\big]$. 
Scores closer to $0$ correspond to more peaked (certain) distributions.

\paragraph{Top Probability}
\[
s_{b,t} = \max_{v} p_{b,t,v} = p_{b,t,(1)}.
\]
A simple peak-confidence heuristic. 
Ignores how close competitors are.

\paragraph{Top Probability Margin}
\[
s_{b,t} = p_{b,t,(1)} - p_{b,t,(2)}.
\]
Measures local ambiguity between the two most likely tokens. 
Larger margin $\Rightarrow$ clearer preference.

\paragraph{Positional Schedule}
\[
s_{b,t} = t.
\]
A deterministic curriculum ignoring model uncertainty (e.g.\ left-to-right). 
Negate or reverse indices if the opposite order is desired.

\subsection{Implementation Details}
Our implementation of PUNT follows the procedure outlined in Algorithm~\ref{alg:planner}. 
To ensure a fair comparison, both PUNT and the baseline methods use the same confidence scoring strategy 
for each model. Specifically, we use the top probability margin for LLaDA and negative entropy for Dream.

\subsubsection*{Sampling and Temperature Settings}
All methods employ nucleus sampling with nucleus mass set to $0.9$. 
We present results for two temperature settings: $0.1$ (low temperature, focused sampling) 
and $0.7$ (higher temperature, more diverse sampling) 
to evaluate robustness across different generation regimes.

\subsubsection*{End-of-Sequence Handling}
To prevent premature termination, we down-weight positions corresponding to end-of-sequence tokens 
when early termination is undesirable:
If $y_{b,t}$ equals a special end-of-sequence token $\text{EOS}$ and early termination is undesirable, enforce
\[
s_{b,t} \leftarrow C_{\text{neg}}, \qquad C_{\text{neg}} \ll 0,
\]
to deprioritize revealing that position.

\subsection{Results and Analysis}
\label{subsec:results}
We present our results grouped by sequence length, as this factor significantly impacts 
the relative performance of the scheduling methods.

\subsubsection*{Short-Sequence Benchmarks} 
We evaluate PUNT on GSM8K, HumanEval, and MBPP
—all tasks with sequences shorter than 1024 tokens (see Table~\ref{tab:config}). 
These benchmarks test mathematical reasoning and code generation capabilities 
under constrained generation lengths.

\textbf{Results:} 
When measured by the number of generation steps, 
PUNT consistently outperforms all baseline methods across both temperature settings (0.1 and 0.7). 
However, when evaluated by NFEs per sequence, PUNT shows competitive but not dominant performance.
PUNT's strength lies in reducing the number of sequential generation steps through aggressive parallelization, 
but each step may require more network evaluations due to its comprehensive independence testing. 

\begin{figure}
  \centering
  \subfloat[GSM8K results at temperature 0.1]{
    \includegraphics[width=\columnwidth]{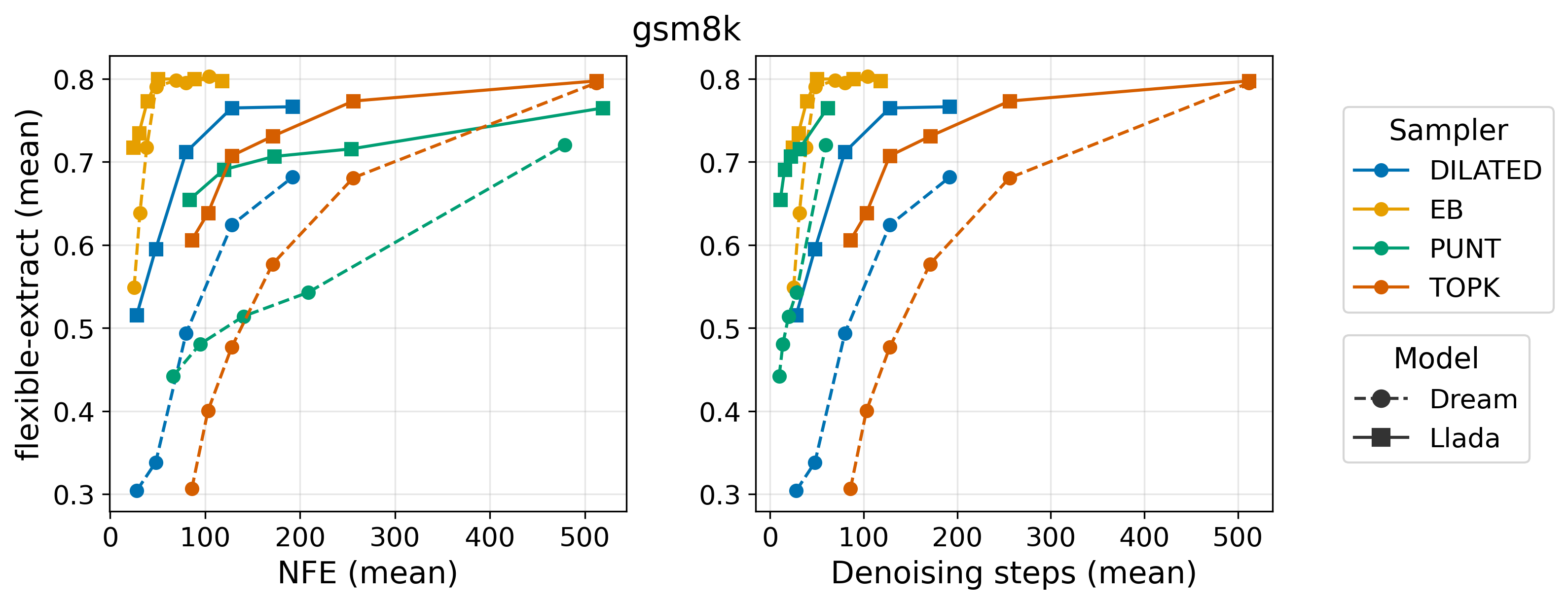}
    \label{fig:gsm8k-temp01}
  }\\
  \subfloat[GSM8K results at temperature 0.7]{
    \includegraphics[width=\columnwidth]{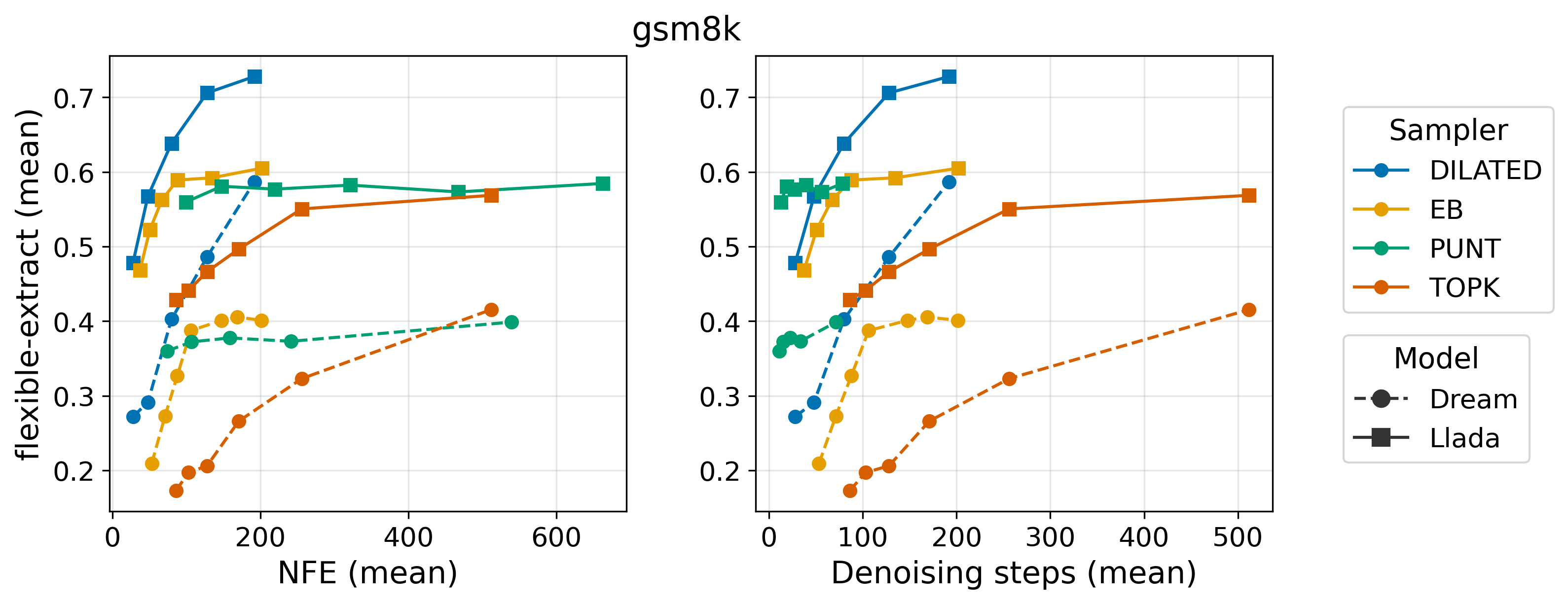}
    \label{fig:gsm8k-temp07}
  }
  \caption{GSM8K performance comparison across different temperature settings showing NFE/steps vs
   match accuracy (flexible-extract filter)}
  \label{fig:gsm8k-combined}
\end{figure}

\begin{figure}
  \centering
  \subfloat[HumanEval results at temperature 0.1]{
    \includegraphics[width=\columnwidth]{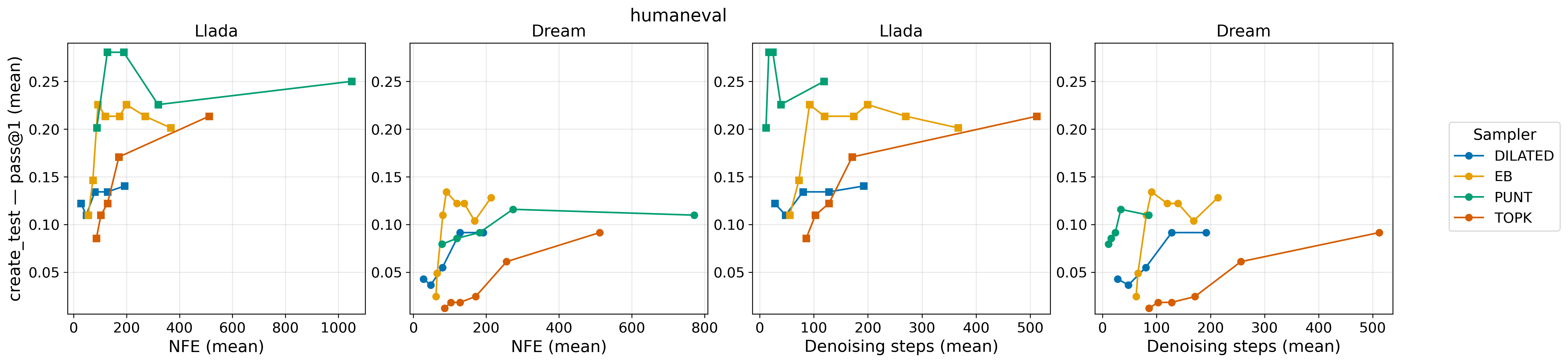}
    \label{fig:humaneval-temp01}
  }\\
  \subfloat[HumanEval results at temperature 0.7]{
    \includegraphics[width=\columnwidth]{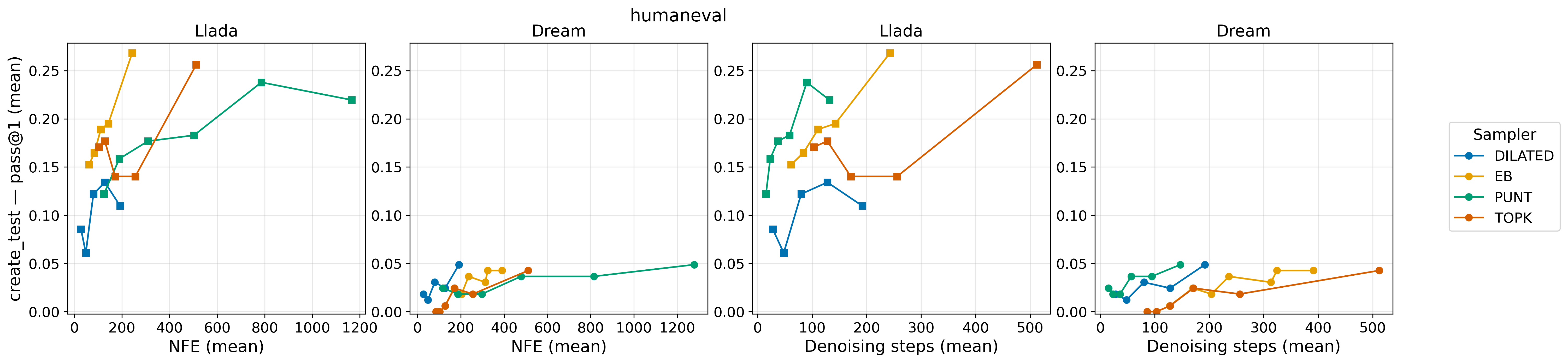}
    \label{fig:humaneval-temp07}
  }
  \caption{HumanEval performance comparison across different temperature settings showing NFE/steps vs Pass@1 success rate for both LLaDA and Dream models}
  \label{fig:humaneval-combined}
\end{figure}

\begin{figure}
  \centering
  \subfloat[MBPP results at temperature 0.1]{
    \includegraphics[width=\columnwidth]{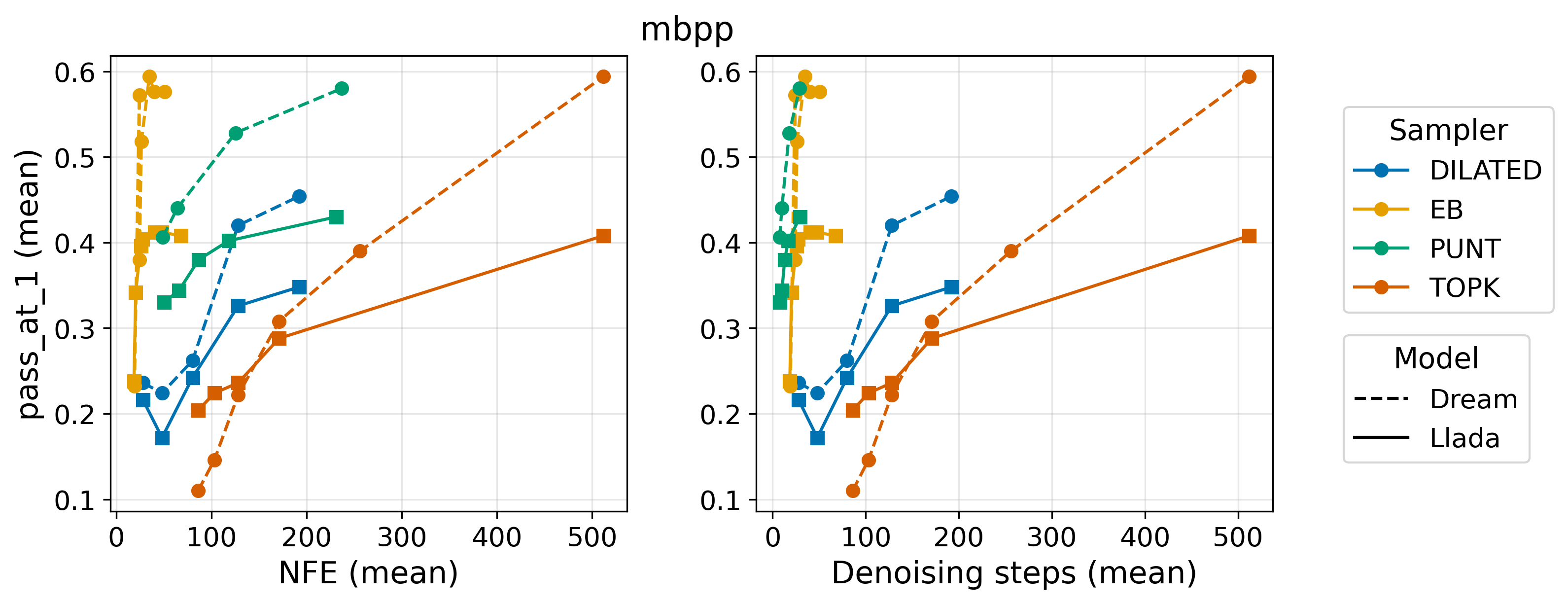}
    \label{fig:mbpp-temp01}
  }\\
  \subfloat[MBPP results at temperature 0.7]{
    \includegraphics[width=\columnwidth]{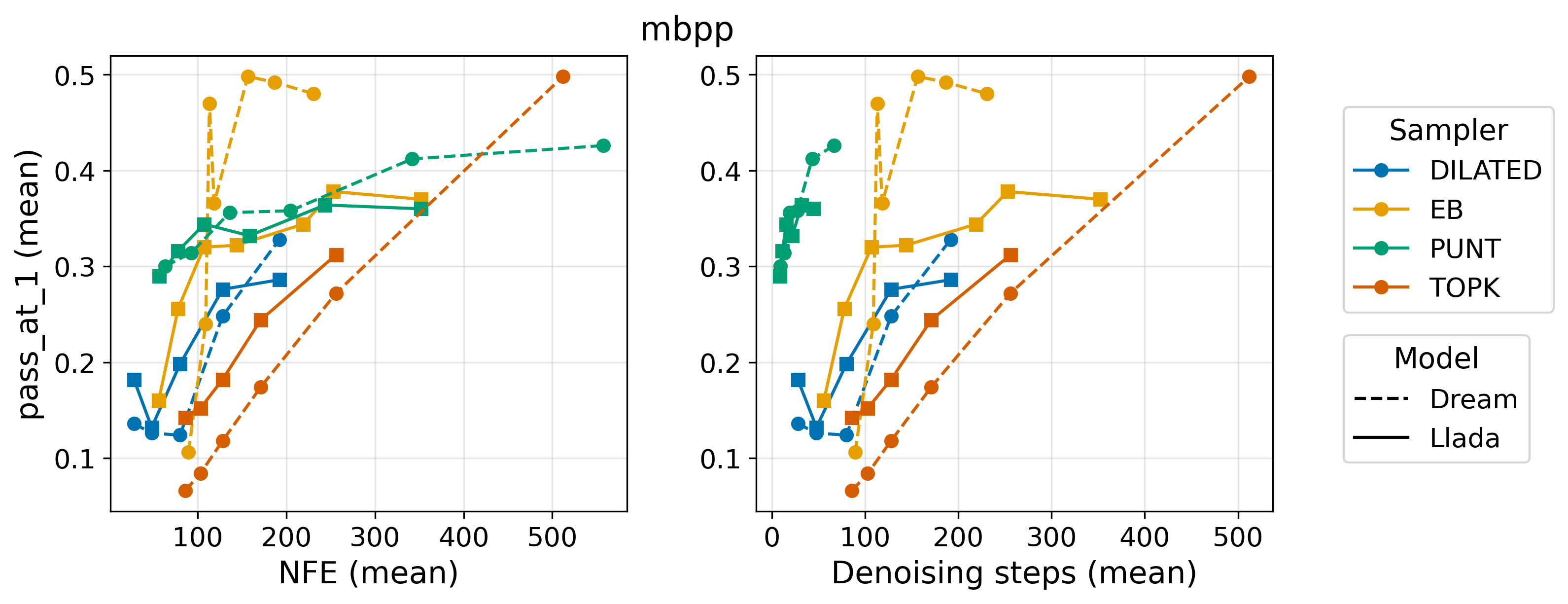}
    \label{fig:mbpp-temp07}
  }
  \caption{MBPP performance comparison across different temperature settings showing NFE/steps vs Pass@1 success rate for both LLaDA and Dream models}
  \label{fig:mbpp-combined}
\end{figure}

\subsubsection*{Long-Sequence Benchmarks} 
For longer sequences ($\geq 1024$ tokens), we evaluate on MT-Bench and IFEval. 
These tasks require sustained coherence and complex instruction following over extended generation windows.

\textbf{MT-Bench Results:} 
MT-Bench consists of open-ended questions spanning creative writing, reasoning, and coding. 
Each question includes two rounds, where the second builds upon the first response. 
Answers are evaluated by GPT-4o using a 1-10 scale. 
All experiments are carried out with temperature 0.7.

\begin{figure}
  \centering
  \subfloat[Dream model results]{
    \includegraphics[width=\columnwidth]{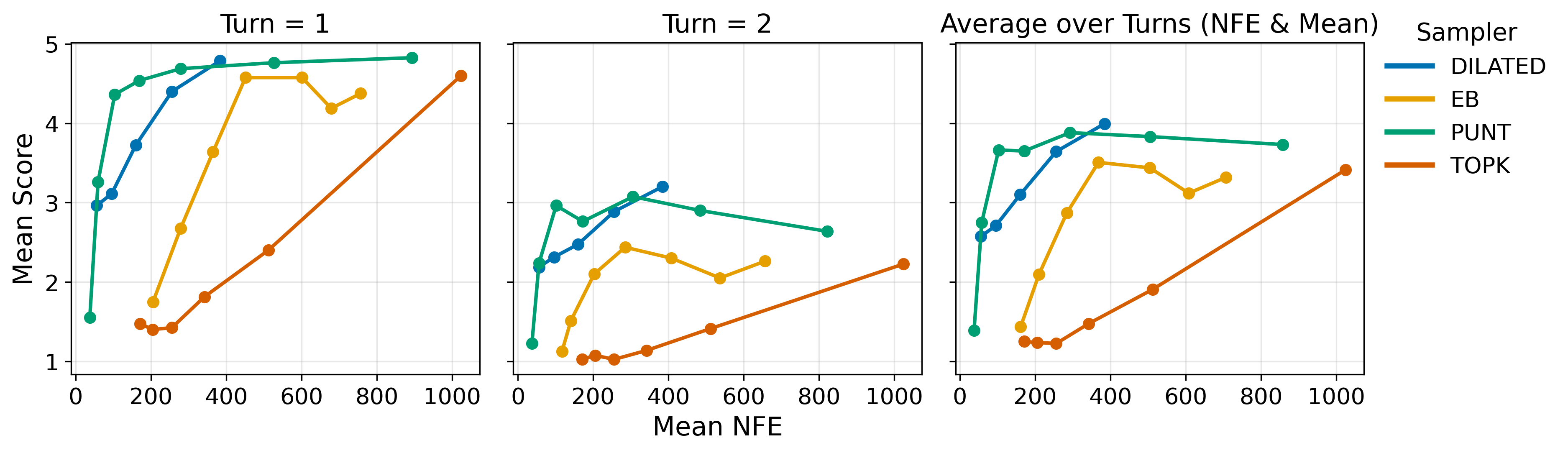}
    \label{fig:dream-results-mtbench}
  }\\
  \subfloat[LLaDA model results]{
    \includegraphics[width=\columnwidth]{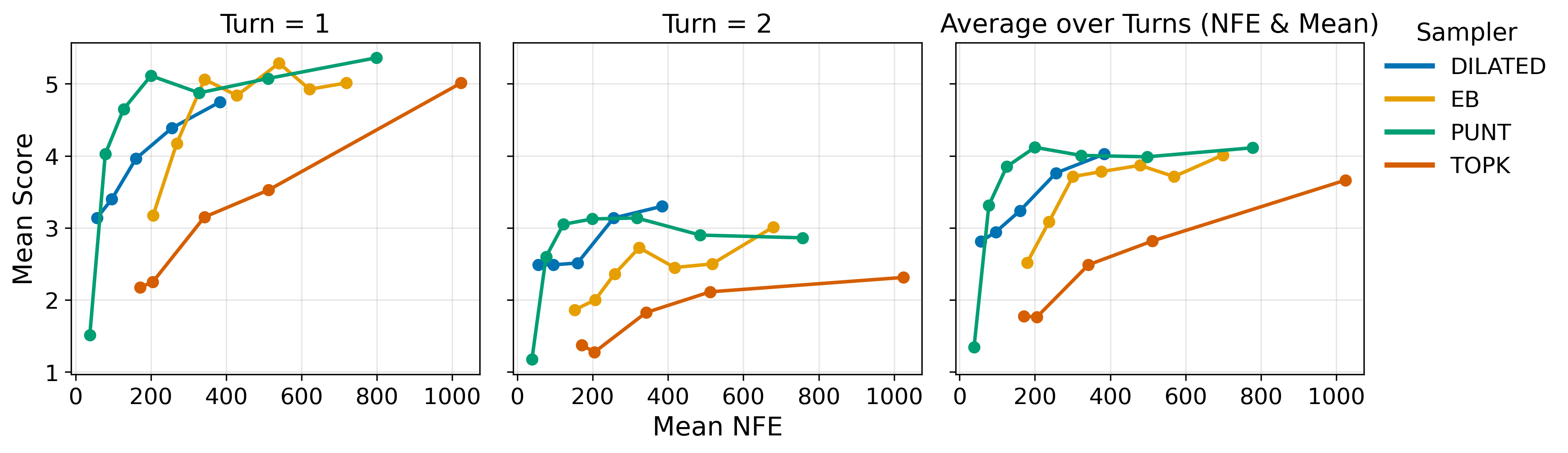}
    \label{fig:llada-results-mtbench}
  }
  \caption{Performance comparison across different models showing NFE vs mean performance}
  \label{fig:combined-results-mtbench}
\end{figure}

\Cref{fig:combined-results-mtbench}  
show that PUNT excels particularly when NFE budgets are severely constrained. 
In low-NFE regimes, PUNT significantly outperforms all baseline methods. 
As the NFE budget increases, dilated sampling begins to show competitive performance, 
but PUNT maintains its characteristic stability advantage.

\textbf{IFEval Results:} 

\begin{figure}
  \centering
  \includegraphics[width=\columnwidth]{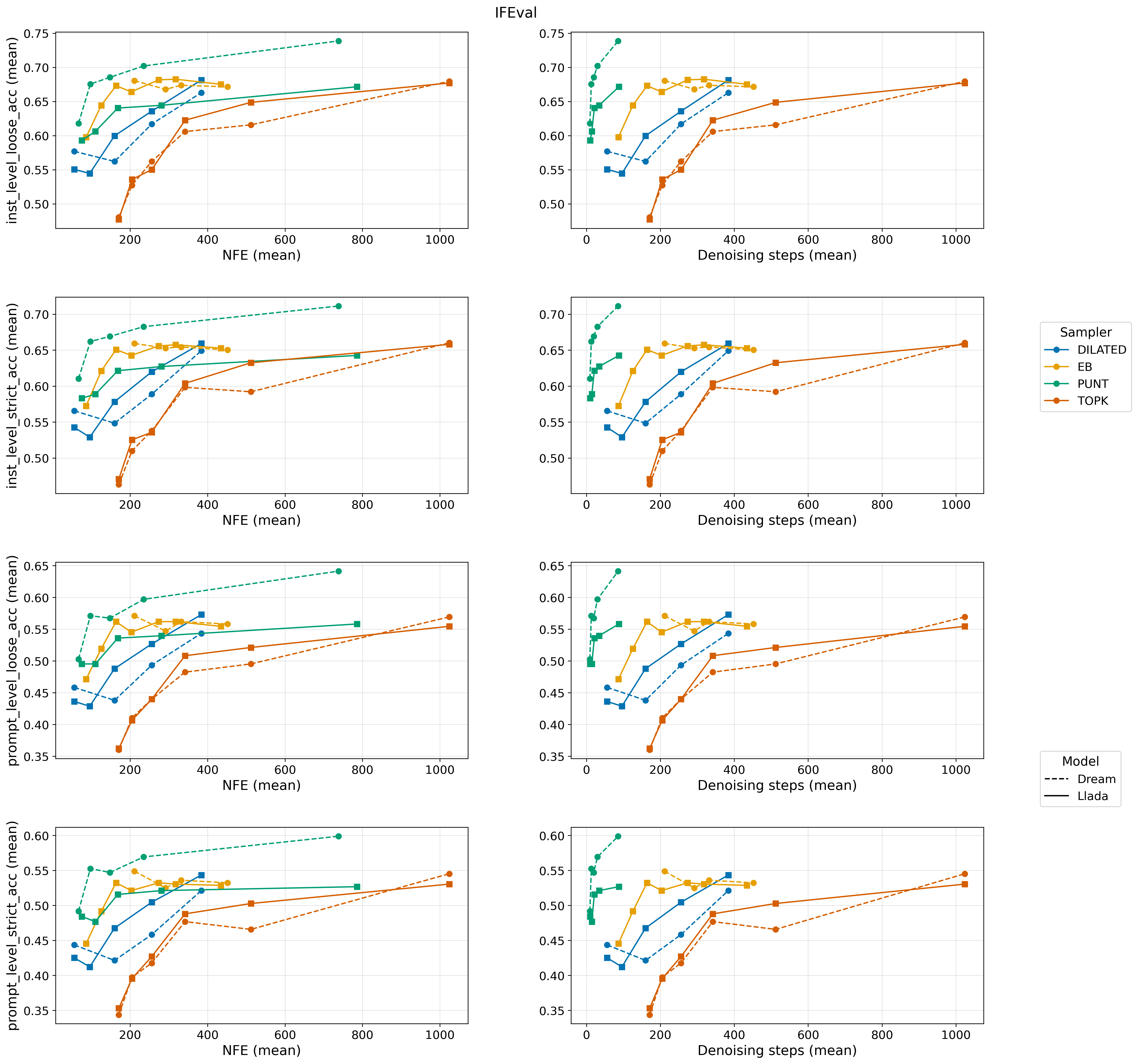}
  \caption{IFEval results showing NFE/steps vs accuracy, temperature 0.1.}
  \label{fig:ifeval-results-temp01}
\end{figure}

\begin{figure}
  \centering
  \includegraphics[width=\columnwidth]{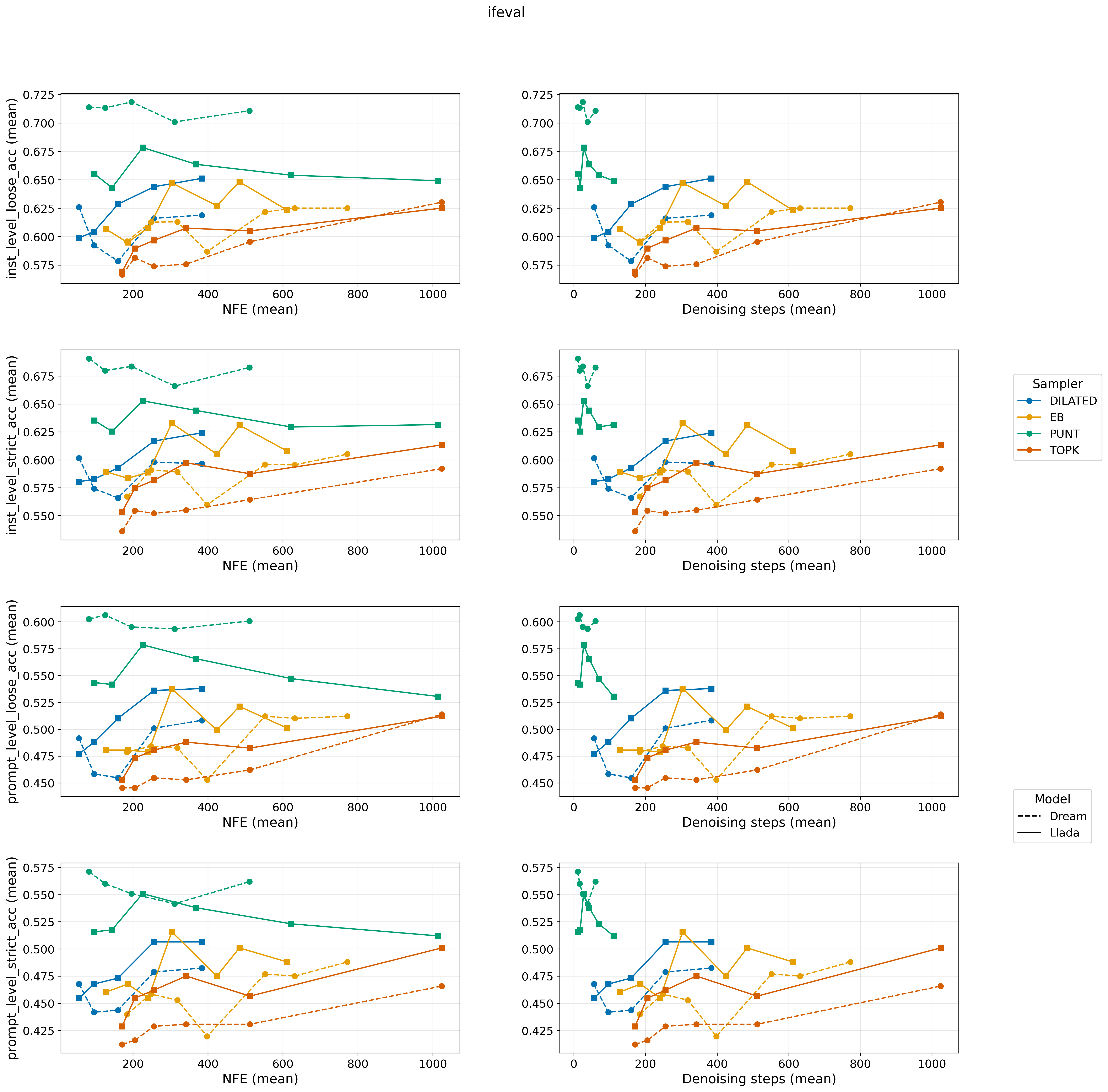}
  \caption{IFEval results showing NFE/steps vs accuracy, temperature 0.7.}
  \label{fig:ifeval-results-temp07}
\end{figure}

The instruction-following evaluation tests adherence to specific formatting and content constraints. 
PUNT demonstrates consistent accuracy across both NFE and step-based metrics, 
again showing its reliability advantage.

PUNT’s results on IFEval demonstrate its stability across different computational budgets. 
As shown in \Cref{fig:ifeval-results-temp01,fig:ifeval-results-temp07}, 
it consistently leads in generation steps at both temperatures, without compromising accuracy. 
Additionally, PUNT is more NFE-efficient at lower budgets and remains competitive as the budget increases, 
pulling ahead at a temperature of 0.7.

\subsection{Masked Diffusion Models for Proteins}
\label{appdx:prot_experiments}

Masked diffusion models (MDMs) have demonstrated effectiveness beyond natural language processing, 
particularly in generating biological sequences such as proteins and DNA. 
To evaluate PUNT's performance in a structured biological domain, 
we conduct experiments on \emph{de novo} membrane protein design using MemDLM~\citep{goel2024memdlm}, 
a masked diffusion model that finetunes the state-of-the-art ESM-2 150M 
protein language model~\citep{lin2023evolutionary} 
with an MDM objective to generate realistic membrane proteins.

\subsubsection{Experimental Setup}
We evaluate PUNT on unconditional protein generation with sequences of up to 1024 amino acids, 
comparing against three established training-free schedulers: 
Top-$k$ sampling, Entropy-Bound (EB) unmasking, and geometry-aware  (Dilated) spacing. 
All methods employ a temperature of 0.8, to encourage sequence novelty, and 
suppress end-of-sequence tokens to promote longer, more realistic protein sequences.
For each sampling strategy, we generate 50 amino acid sequences using the following hyperparameters: 
\begin{itemize}
    \item PUNT: $\varepsilon=\{0.001, 0.004, 0.01, 0.02, 0.04, 0.08, 0.16\}$
    \item Top-$k$: $k=\{1,2,3,4,6,8,12\}$
    \item EB Sampler: $\varepsilon=\{0.1,0.5,1,5,10\}$
    \item Geometry-aware spacing: $\log w = \{3,4,5,6,7,10\}$
\end{itemize}

\subsubsection{Evaluation Metrics}
We assess PUNT's performance across two key dimensions critical for practical 
protein design applications~\citep{wenran2025benchmark}:
\paragraph{Computational Efficiency:} 
As with the natural language benchmarks, 
we measure the number of forward evaluations (NFE) and denoising steps required for generation. 
NFE represents the total number of model forward passes needed to complete sequence generation, 
providing a direct measure of computational cost. Denoising steps (PUNT-specific) 
track the number of iterative refinement steps in the masked diffusion process.

\paragraph{Structural Validity:} 
Generated protein sequences are evaluated for their likelihood to fold into stable, 
well-defined three-dimensional structures. 
We feed each generated amino acid sequence to ESMFold~\citep{lin2023evolutionary} 
to predict the corresponding 3D protein structure. 
We then calculate the mean pLDDT 
(a per-residue measure of local confidence in the structural predictions) 
across all residues in each predicted structure.

\subsubsection{Results and Analysis}

\begin{figure}[h!]
  \centering
  \includegraphics[width=\columnwidth]{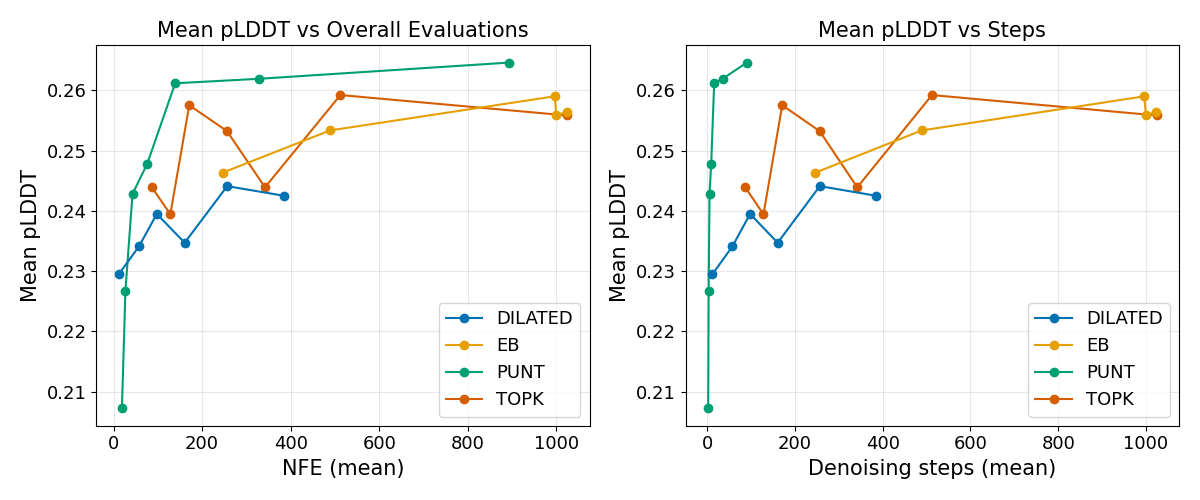}
  \caption{Protein generation with MeMDLM: mean pLDDT vs (a) NFE, and (b) denoising steps.}
  \label{fig:plddt}
\end{figure} 

~\Cref{fig:plddt} plots the mean pLDDT against NFE and number of denoising steps. We find that while the pLDDT of generated structures is low across denoising methods---which may be attributed in part to our use of a non-semi-autoregressive generation strategy, or because of the very long sequence length and absence of multiple sequence alignment in ESMfold making this a challenging domain for 3D structure prediction---PUNT consistently generates proteins with comparable or marginally higher pLDDT than the baseline samplers given the same computational budget, with stable performance across a broad range of NFE.
These results suggest that PUNT is able to improve efficiency without sacrificing structural plausibility, making it well-suited for rapid proposal of candidate proteins for downstream analysis.

\section{Denoising Process}
\label{sec:unmaskings}

In this section, 
we show examples of our denoising process starting from a completely masked response for three
prompts from different domains.

\subsection*{Text Prompt}
The first prompt is a story generation prompt:
\textit{``Compose an engaging story about a recent trip to Hawaii, highlighting 
cultural experiences and must-see attractions."}
\begin{figure}[h]
    \includegraphics[width=\textwidth]{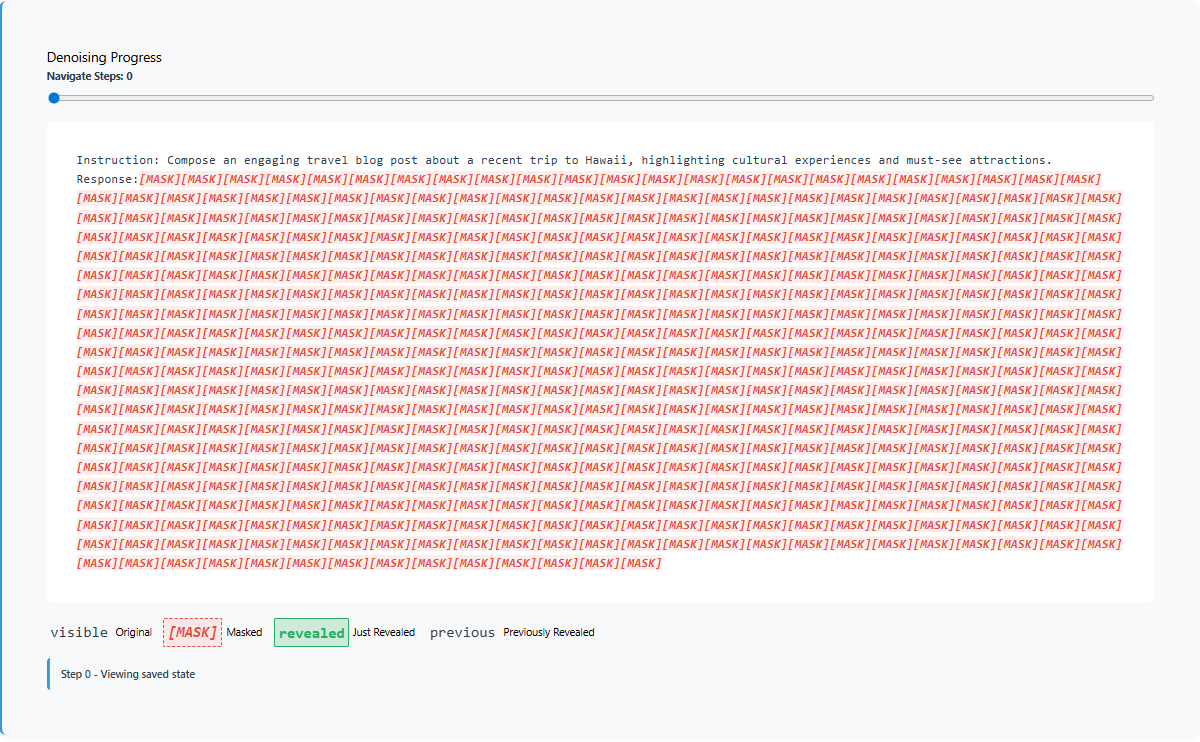}
    \vspace{2mm}
    \includegraphics[width=\textwidth]{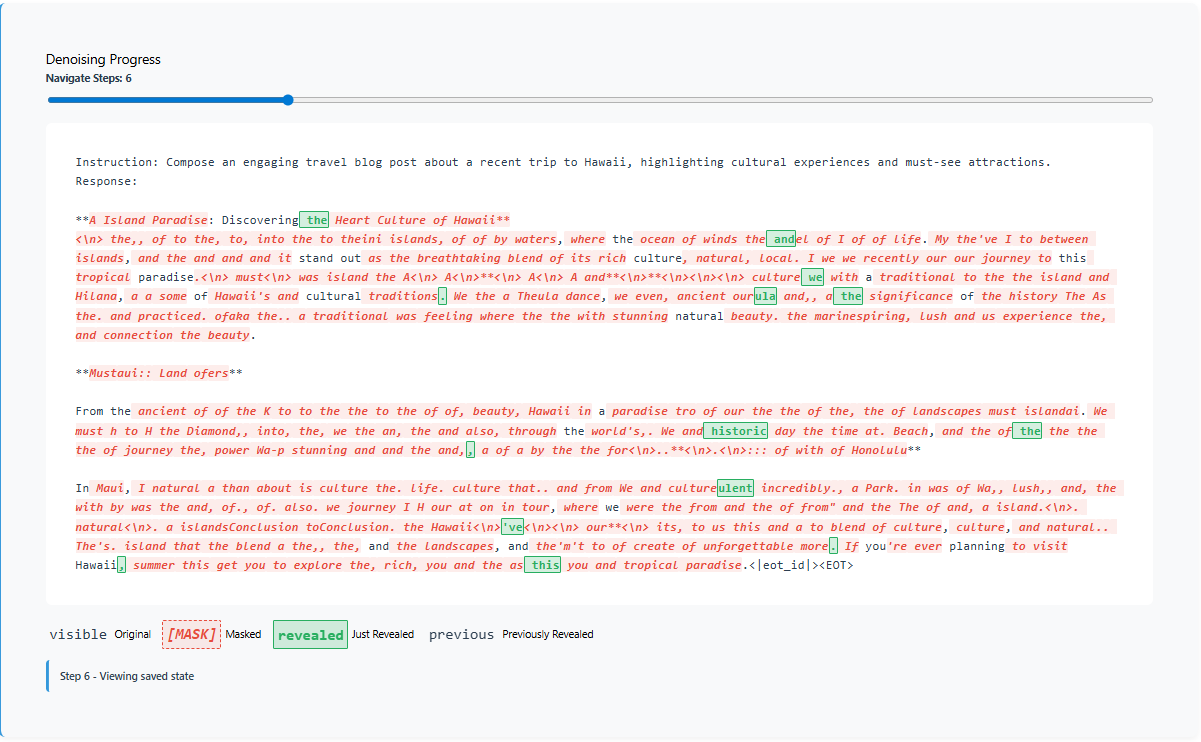}
\end{figure}

\begin{figure}[p]
\ContinuedFloat
\centering
\includegraphics[width=\textwidth]{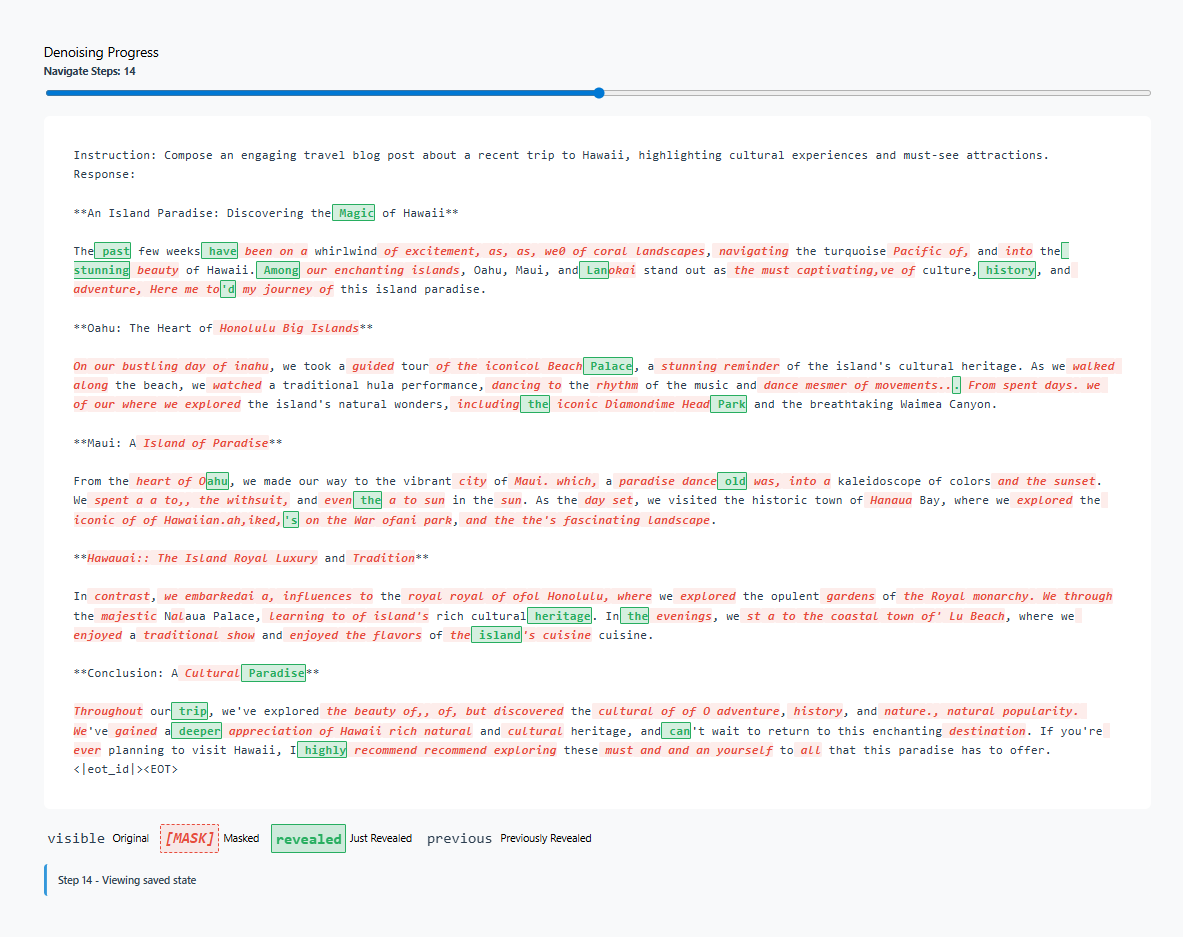}
\vspace{2mm}
\includegraphics[width=\textwidth]{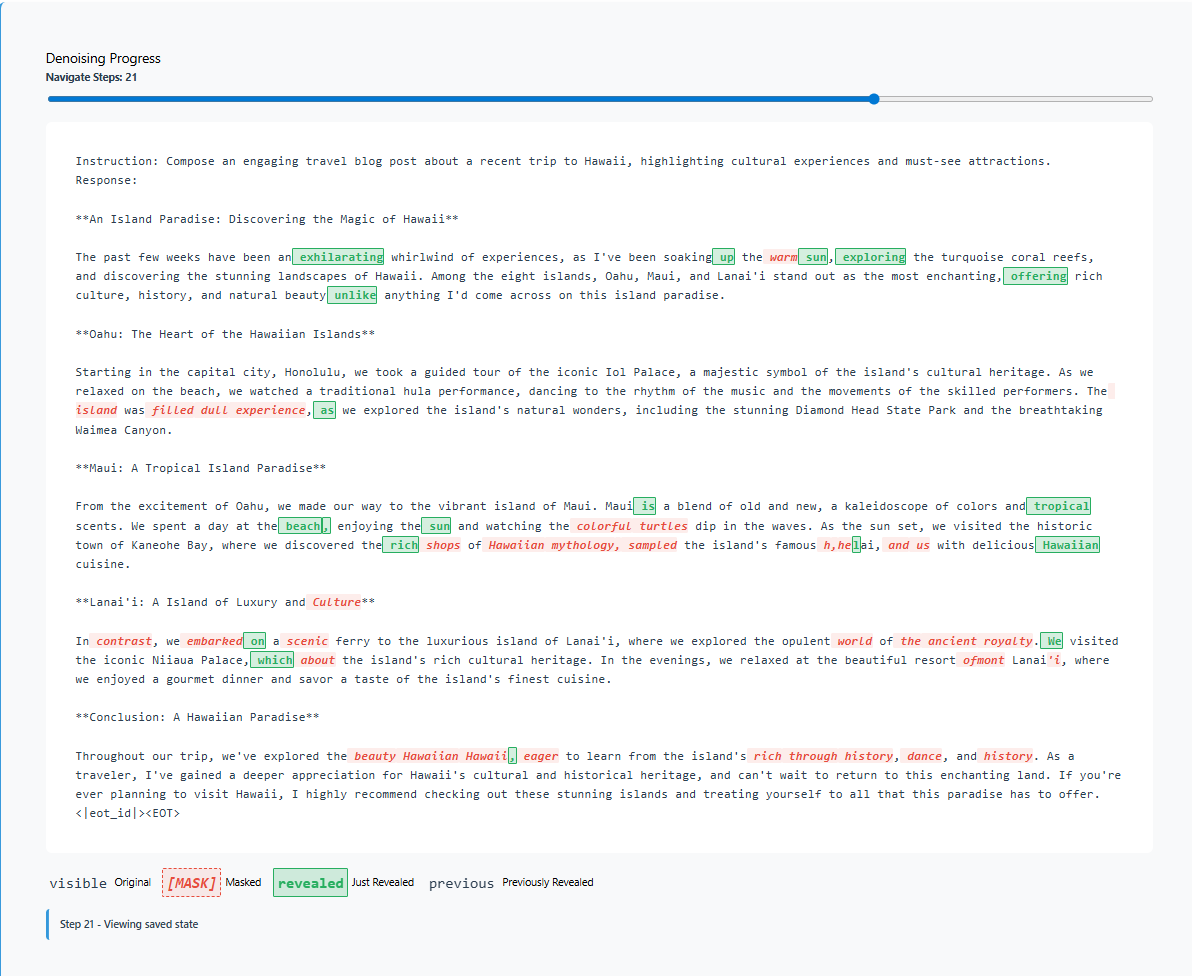}
\end{figure}

\begin{figure}[p]
\ContinuedFloat
\centering
\includegraphics[width=\textwidth]{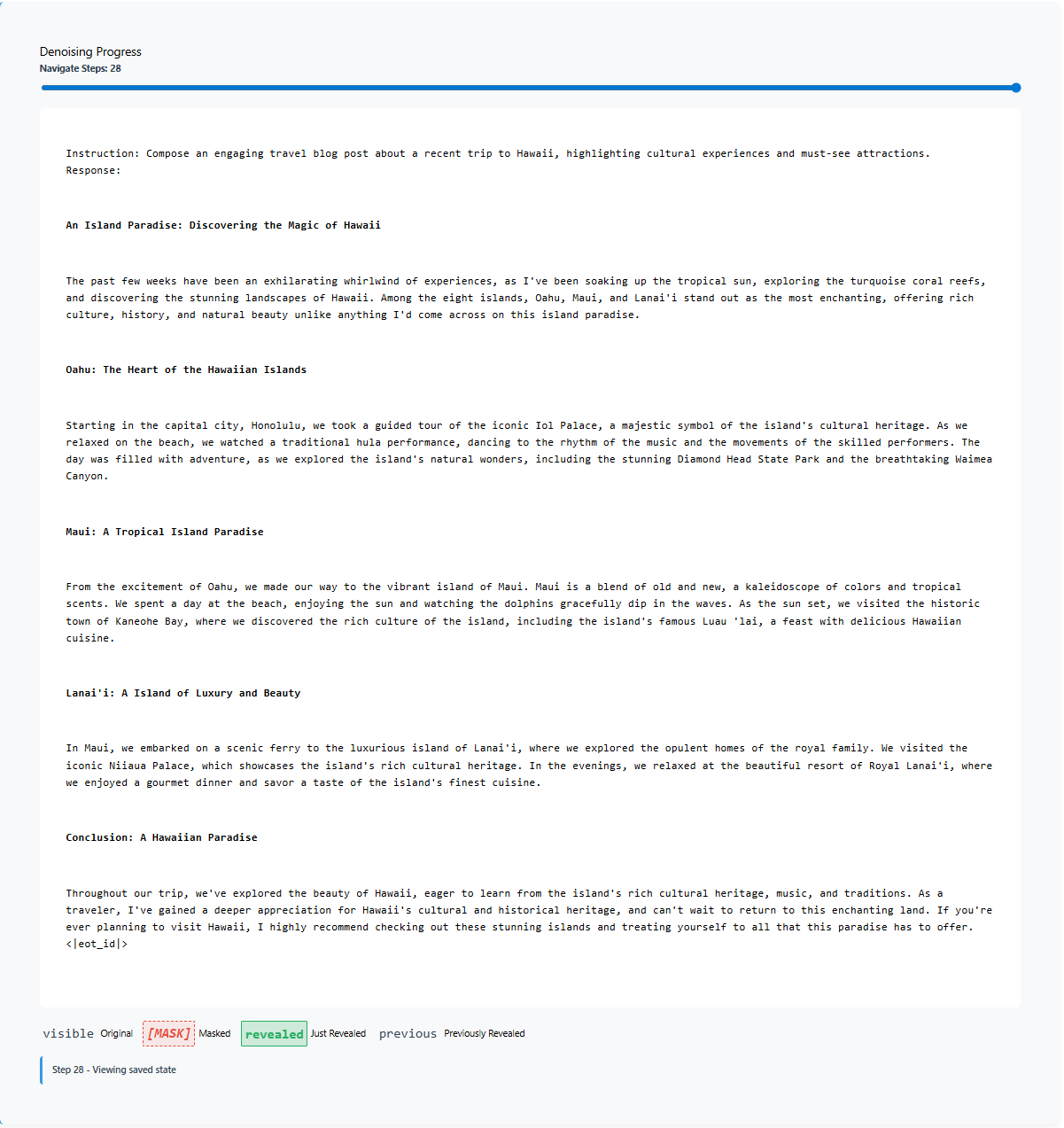}
\end{figure}
\subsection*{Math Prompt}
The second prompt is a math word problem:
\textit{``Natalia sold clips to 48 of her friends in April, and then she sold half 
as many clips in May. How many clips did Natalia sell in altogether in April and May?"}
\begin{figure}[p]
\ContinuedFloat
\centering
\includegraphics[width=\textwidth]{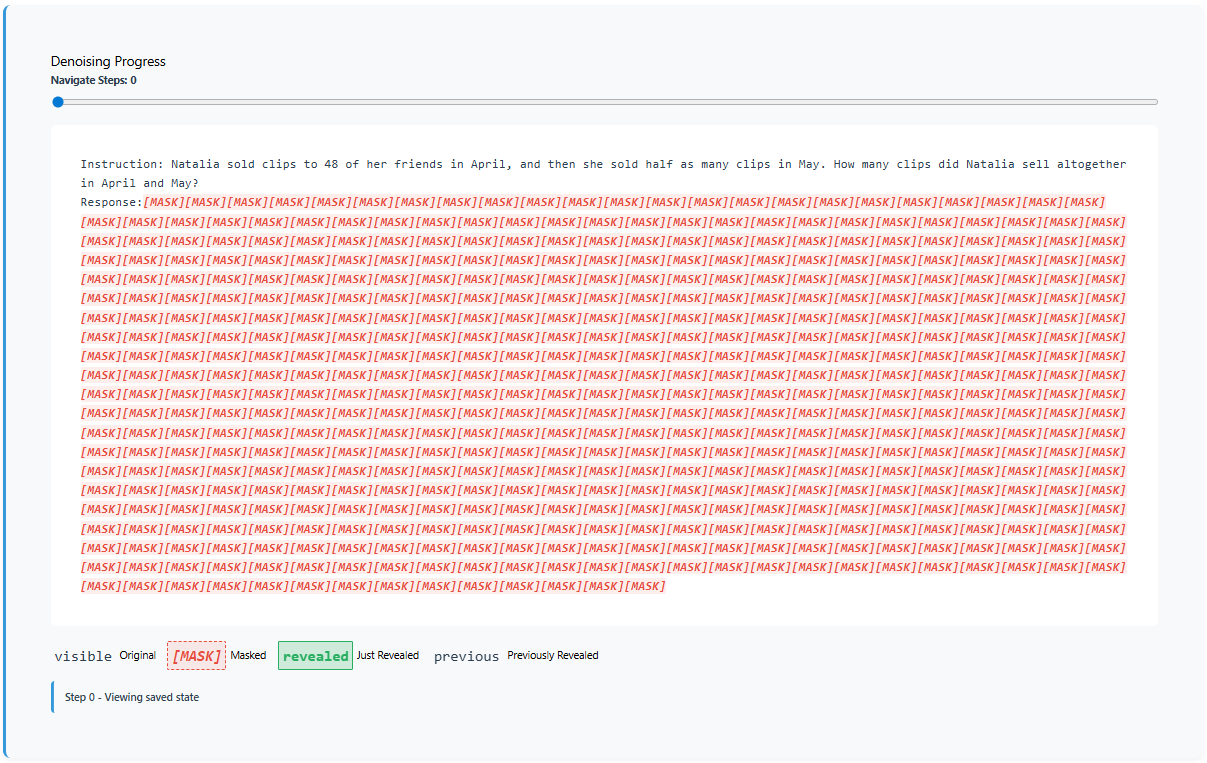}
\vspace{2mm}
\includegraphics[width=\textwidth]{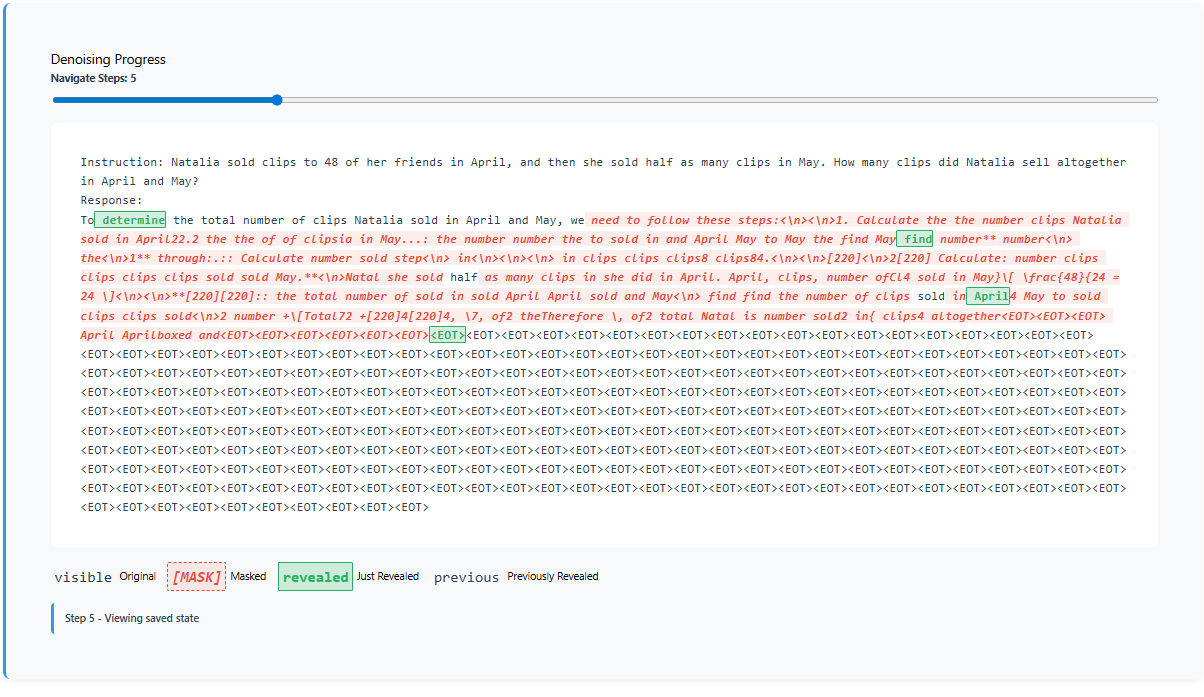}
\vspace{4mm} 
\end{figure}

\begin{figure}[p]
\ContinuedFloat
\centering
\includegraphics[width=\textwidth]{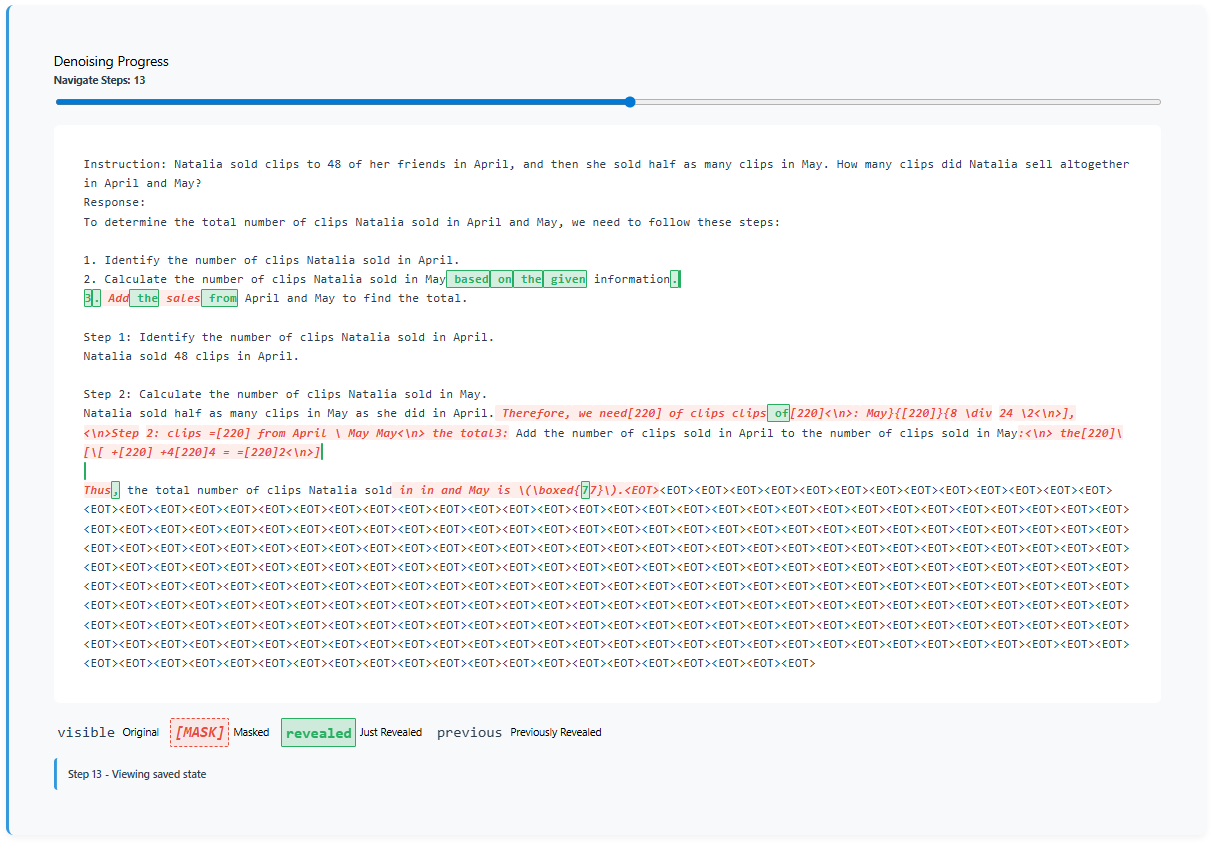}
\vspace{2mm}
\includegraphics[width=\textwidth]{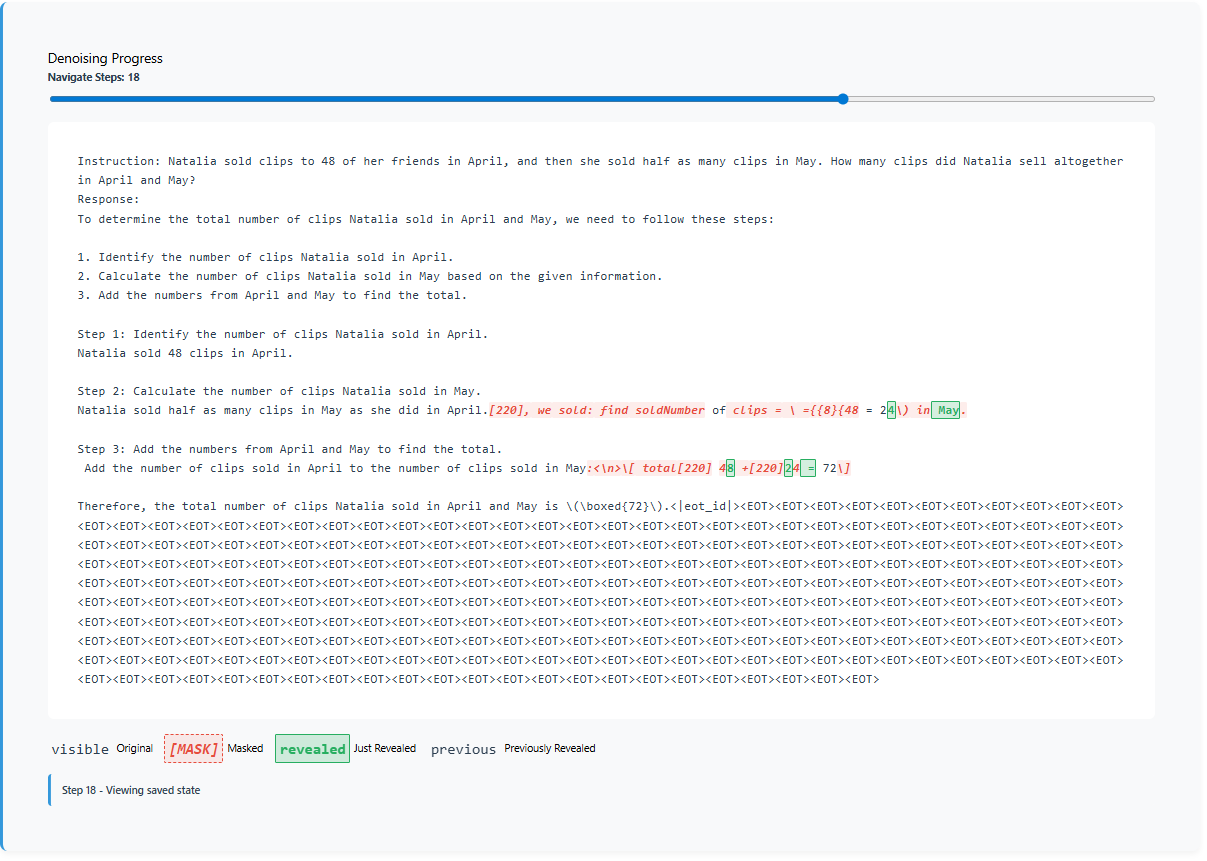}
\vspace{4mm} 
\end{figure}

\begin{figure}[p]
\ContinuedFloat
\centering
\includegraphics[width=\textwidth]{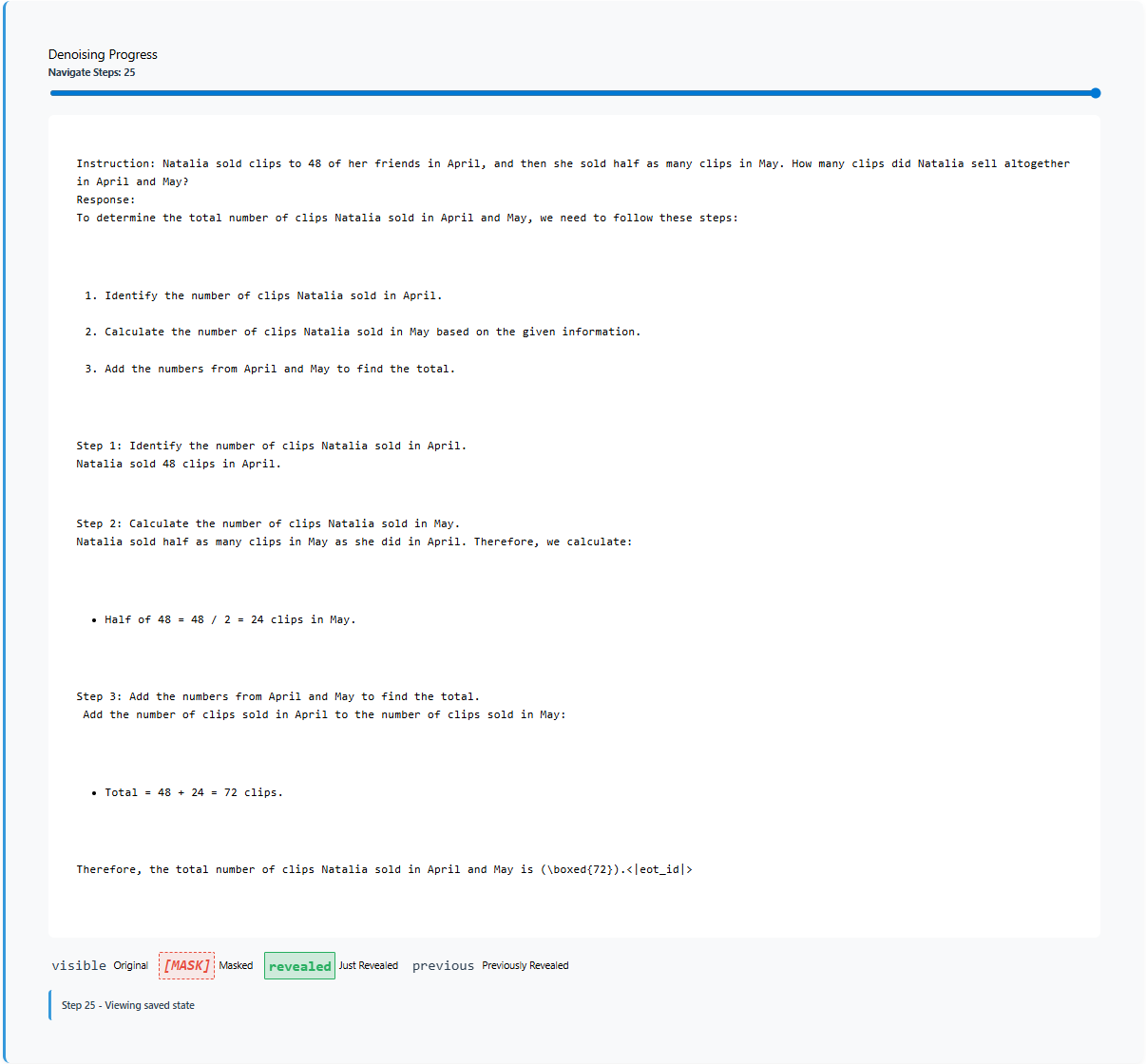}
\end{figure}

\subsection*{Programming Prompt}

The third prompt is a programming prompt:
\textit{``Write a function to find the minimum cost path to reach $(m, n)$ from $(0, 0)$ for the given cost matrix cost$[][]$ and a position $(m, n)$ in cost$[][]$."}

\begin{figure}[h]
\ContinuedFloat
\centering
\includegraphics[width=\textwidth]{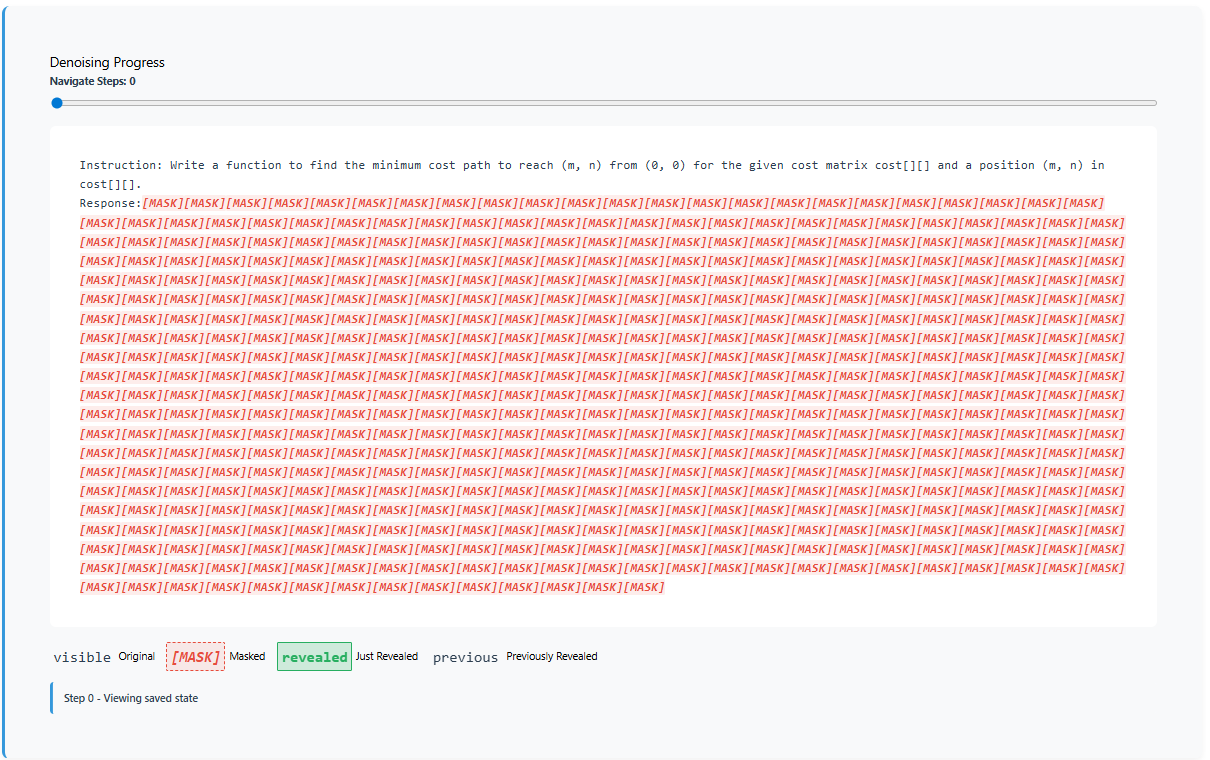}
\vspace{2mm}
\includegraphics[width=\textwidth]{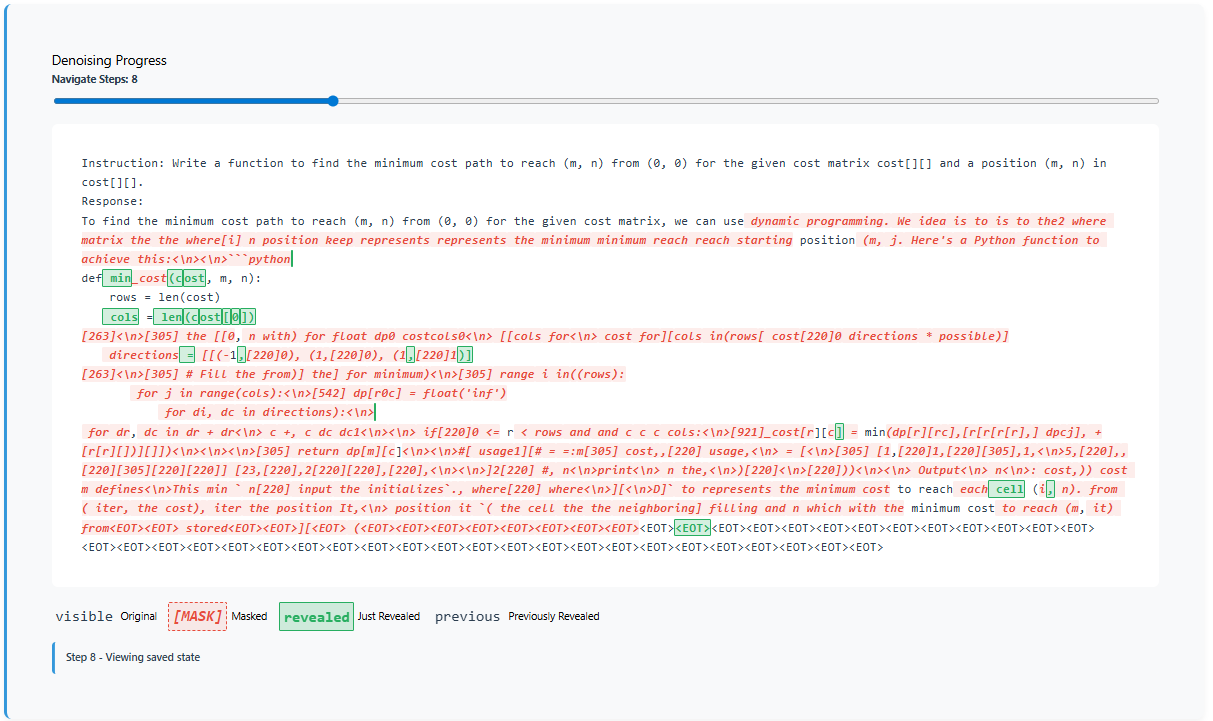}
\vspace{4mm} 
\end{figure}

\begin{figure}[p]
\ContinuedFloat
\centering
\includegraphics[width=\textwidth]{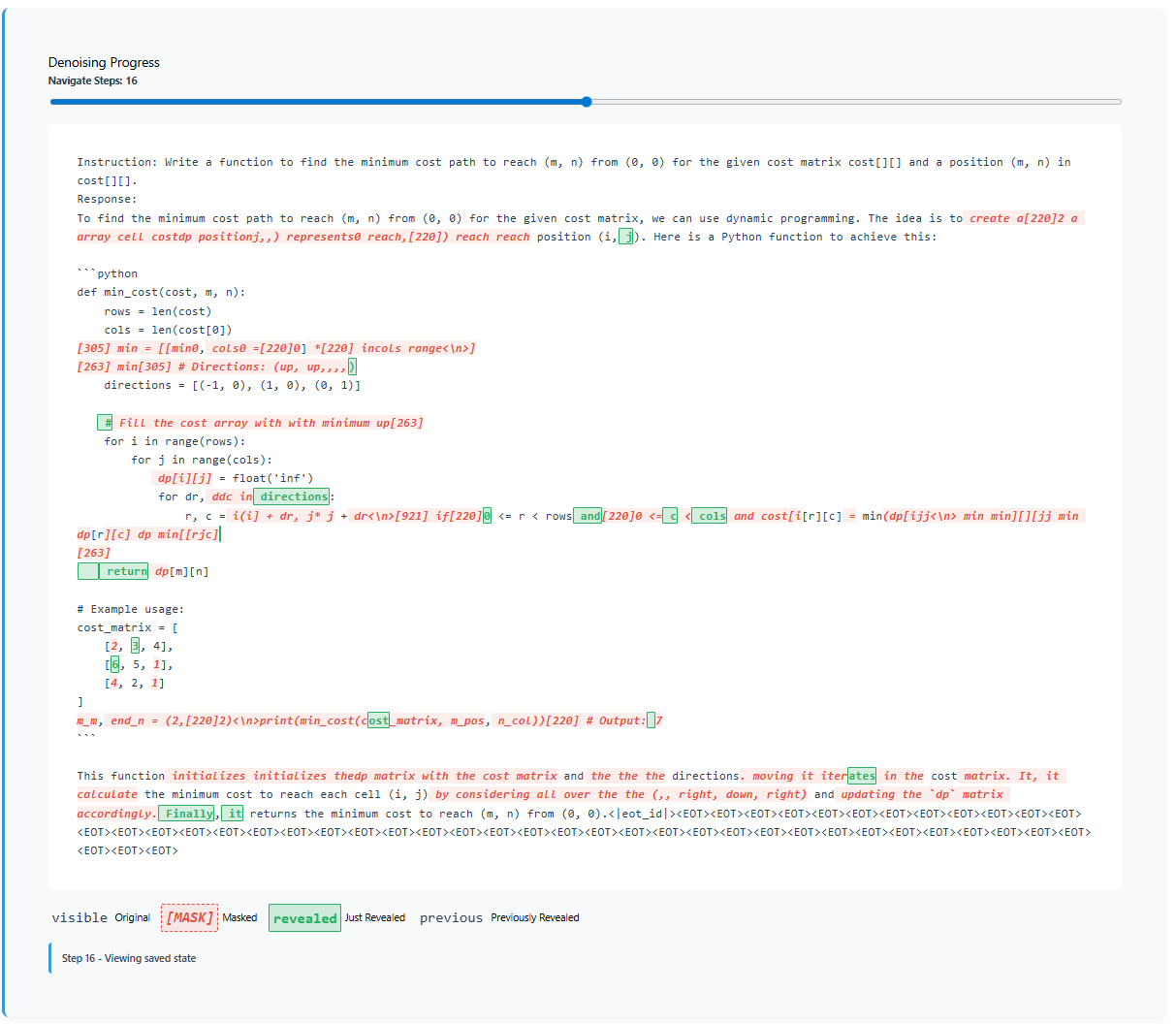}
\vspace{2mm}
\includegraphics[width=\textwidth]{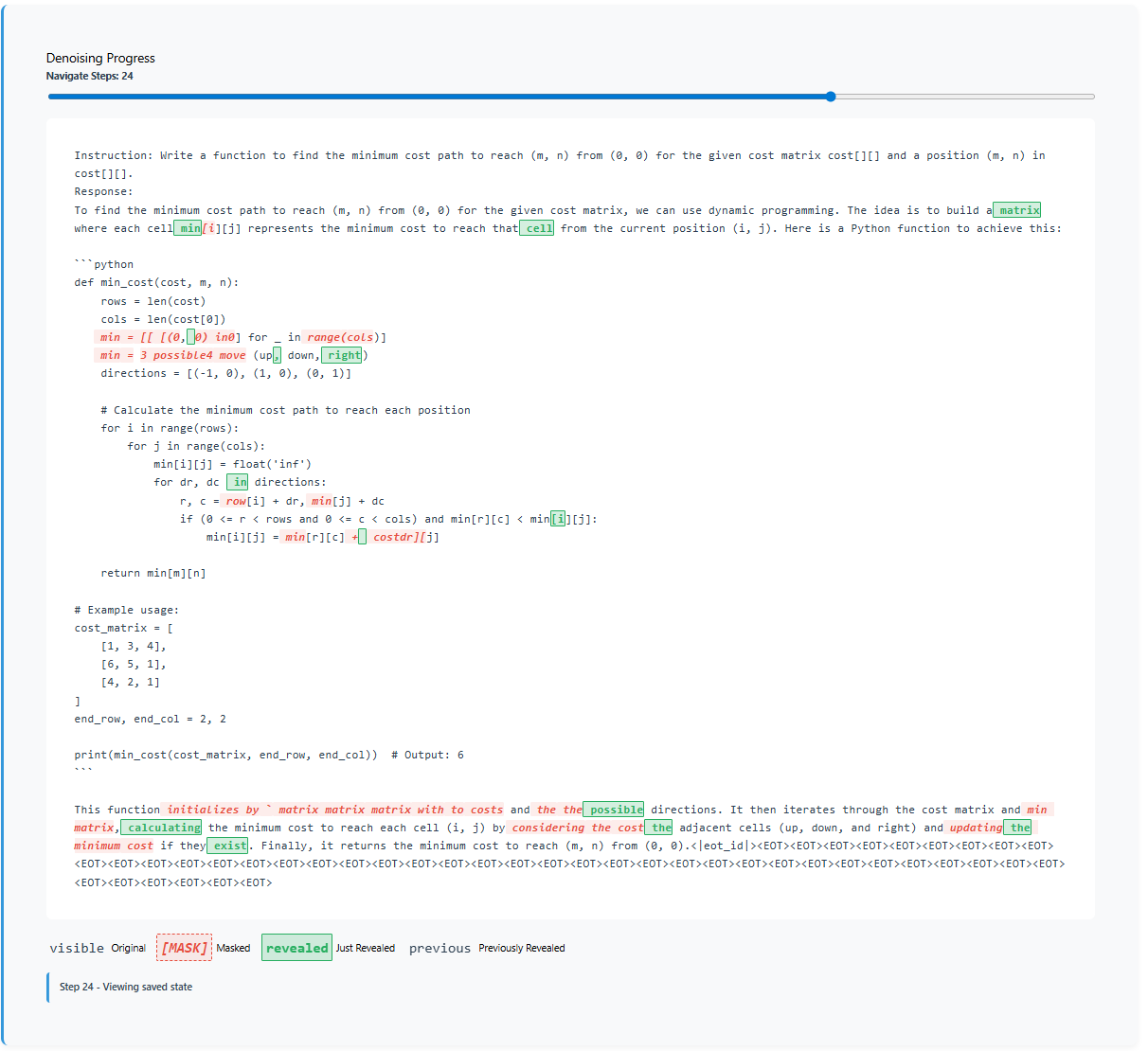}
\vspace{4mm} 
\end{figure}

\begin{figure}[p]
\ContinuedFloat
\centering
\includegraphics[width=\textwidth]{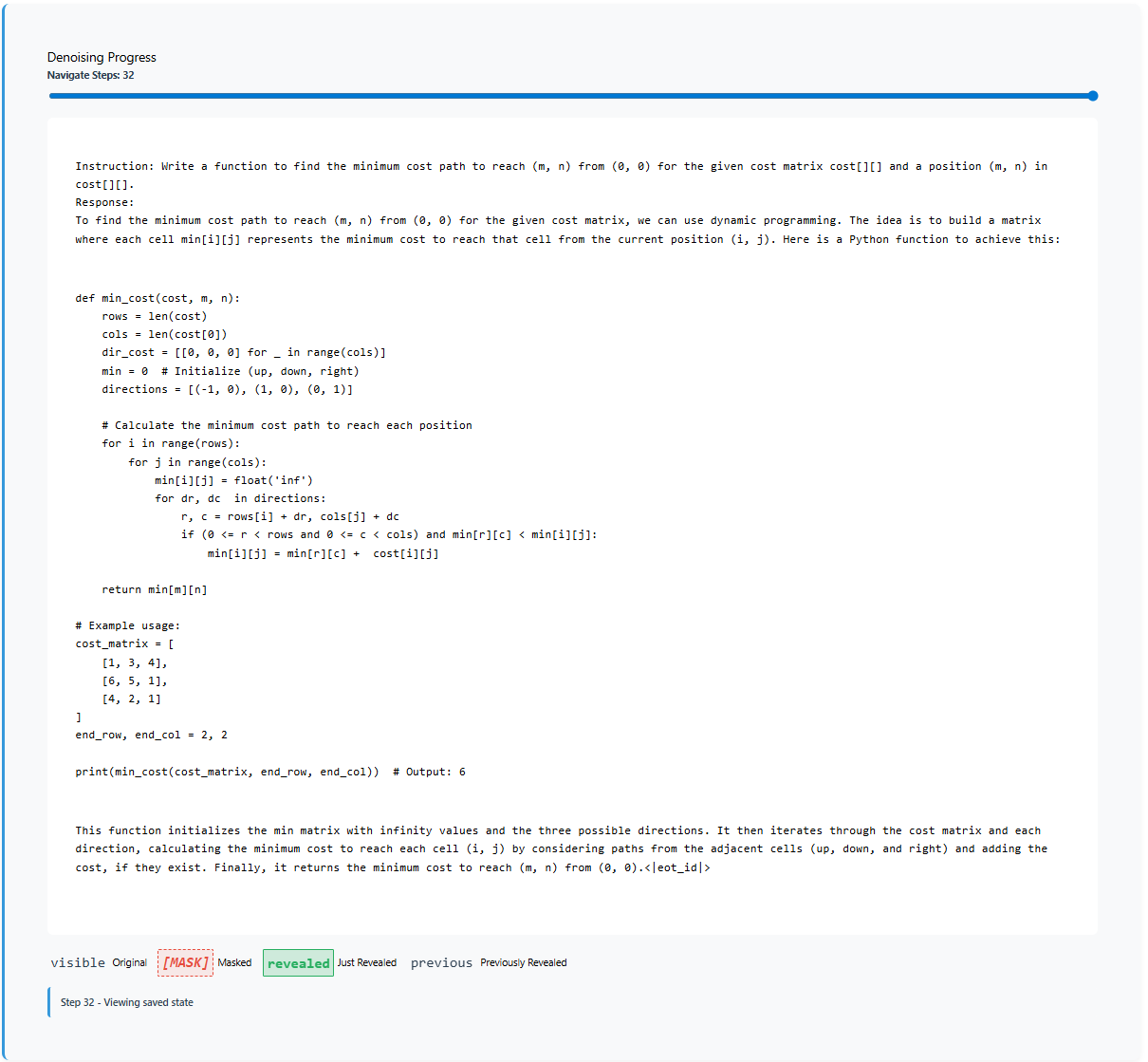}
\end{figure}

\end{document}